\documentclass[10pt,usletter]{article}

\usepackage{amsmath}
\usepackage{amssymb}
\usepackage{amsthm}
\usepackage{times}

\usepackage[totalwidth=450pt,totalheight=640pt]{geometry}

\usepackage{tikz}
\usetikzlibrary{automata}
\usepgflibrary{shapes.geometric}

\newtheorem{DEF}{Definition}
\newtheorem{LEM}{Lemma}
\newtheorem{THE}{Theorem}
\newtheorem{COR}{Corollary}
\newtheorem{CLM}{Claim}

\newtheorem{PRO}{Proposition}

\theoremstyle{definition}
\newtheorem{EXA}{Example}

\newcommand{\SB}{\{\,}%
\newcommand{\SM}{\;{|}\;}%
\newcommand{\SE}{\,\}}%

\newcommand{\Card}[1]{|#1|}
\let\phi=\varphi
\let\epsilon=\varepsilon

\newcommand{\TTT}{\mathcal{T}}
 
\newcommand{\IN}{\textsc{in}}
\newcommand{\OUT}{\textsc{out}}

\renewcommand{\P}{\text{\normalfont P}}
\newcommand{\NP}{\text{\normalfont NP}}
\newcommand{\coNP}{\text{\normalfont co-NP}}

\newcommand{\val}{\text{\normalfont val}}
\newcommand{\ext}{\text{\normalfont ext}}
\newcommand{\GE}{\text{\normalfont GE}}

  \title{\LARGE Algorithms and Complexity Results for Persuasive\\
    Argumentation
    \thanks{Ordyniak and Szeider's research was supported by the
    European Research Council, grant reference~239962.  Kim's research
    was partially supported by the EPSRC, grant reference
    EP/E034985/1.}
  \thanks{A preliminary and shortened version of this paper appeared
    in COMMA 2010.}
  }

  \author{
    \vspace{2mm}
    Eun Jung Kim$^a$, Sebastian Ordyniak$^b$, Stefan Szeider$^b$ \\
    \small $^a$ AlGCo project-team, CNRS, LIRMM, Montpellier, France\\
    \small  $^b$ Vienna University of Technology, Vienna,  Austria
  }

\date{}

\begin{document}


\maketitle
  
\begin{abstract}
  The study of arguments as abstract entities and their interaction as
  introduced by Dung (\emph{Artificial Intelligence} 177, 1995) has
  become one of the most active research branches within Artificial
  Intelligence and Reasoning. A main issue for abstract argumentation
  systems is the selection of acceptable sets of arguments.  Value-based
  argumentation, as introduced by Bench-Capon (\emph{J.~Logic
    Comput.}~13, 2003), extends Dung's framework. It takes into account
  the relative strength of arguments with respect to some ranking
  representing an audience: an argument is subjectively accepted if it
  is accepted with respect to some audience, it is objectively accepted
  if it is accepted with respect to all audiences.

  Deciding whether an argument is subjectively or objectively accepted,
  respectively, are computationally intractable problems.  In fact, the
  problems remain intractable under structural restrictions that render
  the main computational problems for non-value-based argumentation
  systems tractable.  In this paper we identify nontrivial classes of
  value-based argumentation systems for which the acceptance problems
  are polynomial-time tractable.  The classes are defined by means of
  structural restrictions in terms of the underlying graphical structure
  of the value-based system. Furthermore we show that the acceptance
  problems are intractable for two classes of value-based systems that
  where conjectured to be tractable by Dunne (\emph{Artificial
    Intelligence} 171, 2007).
\end{abstract}


\thispagestyle{empty}

\pagestyle{plain}

\section{Introduction}

The study of arguments as abstract entities and their interaction as
introduced by Dung~\cite{Dung95} has become one of the most active
research branches within Artificial Intelligence and Reasoning, see,
e.g.,
\cite{BenchcaponDunne07,BesnardHunter08,RahwanSimari09}. Argumentation
handles possible conflicts between arguments in form of attacks.  The
arguments may either originate from a dialogue between several agents or
from the pieces of information at the disposal of a single agent, this
information may even contain contradictions.  A main issue for any
argumentation system is the selection of acceptable sets of arguments,
where an acceptable set of arguments must be in some sense coherent and
be able to defend itself against all attacking arguments.  Abstract
argumentation provides suitable concepts and formalisms to study,
represent, and process various reasoning problems most prominently in
defeasible reasoning (see, e.g., \cite{Pollock92},
\cite{BondarenkoDungKowalskiToni97}) and agent interaction (see, e.g.,
\cite{ParsonsWooldridgeAmgoud03}).


Extending Dung's concept, Bench-Capon~\cite{BenchCapon03} introduced
\emph{value-based argumentation} systems that allow to compare arguments
with respect to their relative strength such that an argument cannot
successfully attack another argument that is considered of a higher
rank. The ranking is specified by the combination of an assignment of
\emph{values} to arguments and an ordering of the values; the latter is
called an \emph{audience}~\cite{BenchcaponDoutreDunne07}.  As laid out
by Bench-Capon, the role of arguments in this setting is to persuade
rather than to prove, demonstrate or refute. Whether an argument can be
accepted with respect to \emph{all possible} or \emph{at least one}
audience allows to formalize the notions of \emph{objective acceptance}
and \emph{subjective acceptance}, respectively.

An important limitation for using value-based argumentation systems in
real-world applications is the \emph{computational intractability} of
the two basic acceptance problems: deciding whether a given argument is
subjectively accepted is $\NP$-hard, deciding whether it is objectively
accepted is $\coNP$-hard \cite{DunneBenchcapon04}.  Therefore it is
important to identify classes of value-based systems that are still
useful and expressible but allow a polynomial-time tractable acceptance
decision.  However, no non-trivial tractable classes of value-based
systems have been identified so far, except for systems with a tree
structure where the degree of nodes and the number of nodes of degree
exceeding~2 are bounded~\cite{Dunne10}.  In fact, as pointed out by
Dunne~\cite{Dunne07}, the acceptance problems remain intractable for
value-based systems whose graphical structures form trees, in strong
contrast to the main computational problems for non-value-based
argumentation that are linear-time tractable for tree systems, or more
generally, for systems of bounded treewidth~\cite{Dunne07} .

\paragraph{Our Contribution} In this paper we introduce nontrivial
classes of value-based systems for which the acceptance problems are
tractable. The classes are defined in terms of the following notions:
\begin{itemize}
\item The \emph{value-width} of a value-based system is the largest number of
arguments of the same value.
\item The \emph{extended graph structure} of a value-based system has
  as nodes the arguments of the value-based system, two arguments are
  joined by an edge if either one attacks the other or both share the
  same value.
\item The \emph{value graph} of a value-based system has as vertices the
  values of the system, two values $v_1$ and $v_2$ are
  joined by a directed edge if some argument of value $v_1$ attacks an
  argument of value $v_2$~\cite{Dunne10}.
\end{itemize}

We show that the acceptance problems are
tractable for the following classes of value-based systems:
\begin{itemize}
\item[(P1)] value-based systems with a bipartite graph structure where at most
  two arguments share the same value (i.e., systems of
  value-width~2); 
\item[(P2)] value-based systems whose extended graph structure has
  bounded treewidth; and
\item[(P3)] value-based systems of bounded value-width whose value
  graphs have bounded treewidth.
\end{itemize}
In fact, we show that both acceptance problems are \emph{linear time
  tractable} for the classes (P2) and~(P3), the latter being a subclass
of the former.  Our results suggest that the extended graph structure is
a suitable structural representation of value-based argumentation
systems. The positive results (P1)--(P3) hold for systems with
unbounded number of arguments, attacks and values.

We contrast our positive results with negative results that rule out
classes conjectured to be tractable.  We show that the acceptance
problems are (co)-NP-hard for the following classes:
\begin{itemize}
\item[(N1)] value-based systems of value-width 2;
\item[(N2)] value-based systems where the number of attacks between
  arguments of the same value is bounded (systems of \emph{bounded
    attack-width});
\item[(N3)] value-based systems with bipartite value graphs.
\end{itemize}
In fact, we show that both acceptance problems are intractable for
value-based systems of value-width 2 and attack-width~1.  Classes (N1)
and (N2) were conjectured to be tractable~\cite{Dunne07}, the complexity
of (N3) was stated as an open problem~\cite{Dunne10}.

\bigskip\noindent The reminder of the paper is organized as follows.  In
Section~\ref{sec:pre} we provide basic definitions and preliminaries.
In Section~\ref{sec:valuewidth} we define the parameters value-width and
attack-width and establish the results involving systems of
value-width~2, we also discuss the relationship between systems of
value-width 2 and dialogues~\cite{BenchcaponDoutreDunne07}.  In
Section~\ref{sec:tw} we consider value-based systems with an extended
graph structure of bounded treewidth and show linear time tractability.
We close in Section~\ref{sec:concl} with concluding remarks.  Some
proofs of technical lemmas are given in an appendix.

The main results of this paper have been presented in preliminary and
shortened form at COMMA'10~\cite{KimOrdyniakSzeider10}. Here we provide
full proofs, examples, and additional discussions.  Further new
additions are the results (P3) and (N3) involving value graphs, and the
discussion of the relationship between systems of value-width 2 and
dialogues.

\section{Arguments, attacks, values, and audiences}\label{sec:pre}

In this section we introduce the objects of our study more formally.

\subsection{Abstract argumentation system}

\begin{DEF}
  An \emph{abstract argumentation system} or \emph{argumentation
    framework} is a pair $(X,A)$ where $X$ is a finite set of elements
  called \emph{arguments} and $A\subseteq X\times X$ is a binary
  relation called the \emph{attack relation}. If $(x,y)\in A$ we say
  that \emph{$x$ attacks $y$}.
\end{DEF}

An abstract argumentation system $F=(X,A)$ can be considered as a
directed graph, and therefore it is convenient to borrow notions and
notation from the theory of directed graphs \cite{BangjensenGutin00}.
For example we say that a system $F=(X,A)$ is \emph{acyclic} if $(X,A)$
is a DAG (a directed acyclic graph).

\begin{figure}[tbh]
\centering
  \begin{tikzpicture}[scale=0.6]
    \small
    \tikzstyle{every node}=[circle,minimum size=6mm,inner sep=0pt]
    \draw
    (0,0)   node[draw] (A) {$a$} 
    (1.5,0) node[draw] (B) {$b$} 
    (0,-2)   node[draw] (C) {$c$} 
    (1.5,-2) node[draw] (D) {$d$} 
    (1.5,-4) node[draw] (F) {$f$} 
    (0,-4)   node[draw] (E) {$e$}
    (.75,-5.5) node () {$F_0$} 
    ;
    \draw[thick,->,shorten >=1pt,>=stealth]
    (B) edge (A)
    (C) edge (D)
    (D) edge (B)
    (A) edge (D)
    (F) edge (C)
    (A) edge[bend right] (E)
    ;
  \end{tikzpicture}%
  \hspace{2cm}
  \begin{tikzpicture}[scale=0.6]
    \small
    \tikzstyle{every node}=[circle,minimum size=6mm,inner sep=0pt]
    \draw
    (0,0)   node[draw] (A) {$a$}  
    (1.5,0) node[draw] (B) {$b$} 
    (0,-2)   node[draw] (C) {$c$} 
    (1.5,-2) node[draw] (D) {$d$} 
    (1.5,-4) node[draw] (F) {$f$} 
    (0,-4)   node[draw] (E) {$e$}
    (.75,-5.5) node () {$F$} 
    ;
    \draw[thick,->,shorten >=1pt,>=stealth]

    (B) edge (A)
    (C) edge (D)
    (D) edge (B)
    (A) edge (D)
    (F) edge (C)
    (A) edge[bend right] (E)
    ;
    \draw  (0.75,0) ellipse (1.95 and 0.85);
    \draw  (0.75,-2) ellipse (1.95 and 0.85);
    \draw  (0.75,-4) ellipse (1.95 and 0.85);
    \draw  (-1.6,0) node () {$S$:} ;
    \draw  (-1.6,-2) node () {$E$:} ;
    \draw  (-1.6,-4) node () {$T$:} ;
  \end{tikzpicture}
\caption{ The abstract argumentation system $F_0$ and
  value-based system $F$ of Examples~\ref{exa:af} and \ref{exa:vaf},
  respectively.}\label{fig:af}
\end{figure}
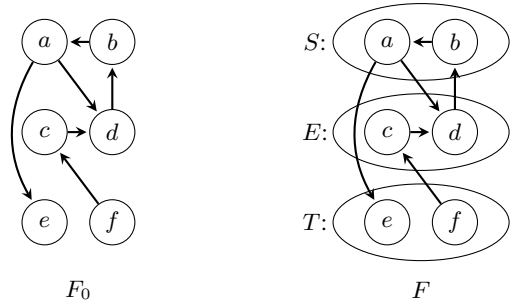

\begin{EXA}\label{exa:af}
  An abstract argumentation system $F_0=(X,A)$ with arguments $X=\{a$,
  $b$, $c$, $d$, $e$, $f\}$ and attacks $A=\{(a,d)$, $(a,e)$, $(b,a)$,
  $(c,d)$, $(d,b)$, $(f,c)\}$ is displayed in Figure~\ref{fig:af}.
\end{EXA}
Next we define commonly used semantics of abstract argumentation systems
as introduced by Dung~\cite{Dung95}. For the discussion of other
semantics and variants, see, e.g., Baroni and Giacomin's
survey~\cite{BaroniGiacomin09}.  Let $F=(X,A)$ be an abstract
argumentation system and $S\subseteq X$.
\begin{enumerate}
\item $S$ is \emph{conflict-free} in $F$  if there is no $(x,y)\in A$
  with $x,y\in S$.
\item $S$ is \emph{acceptable} in $F$  if for each
  $x\in S$ and each $y\in X$ with $(y,x)\in A$ there is some $x'\in S$
  with $(x',y)\in A$.
\item $S$ is \emph{admissible} in $F$  if it is conflict-free and acceptable.
\item $S$ is a \emph{preferred extension} of $F$ if $S$ is admissible in $F$
  and there is no admissible set~$S'$ of~$F$ that properly contains $S$.
\end{enumerate}
For instance, the admissible sets of the abstract argumentation system
$F_0$ of Example~\ref{exa:af} are the sets $\emptyset$ and $\{f\}$,
hence $\{f\}$ is its only preferred extension.

Let $F=(X,A)$ be an abstract argumentation system and $x_1\in X$. The
argument $x_1$ is \emph{credulously accepted} in~$F$ if $x_1$ is
contained in some preferred extension of $F$, and $x_1$ is
\emph{skeptically accepted} in $F$ if $x_1$ is contained in all
preferred extensions of~$F$.

In this paper we are especially interested in finding preferred
extensions in \emph{acyclic} abstract argumentation systems. It is well
known that every acyclic system $F=(X,A)$ has a unique preferred
extension $\GE(F)$, and that $\GE(F)$ can be found in polynomial time
($\GE(F)$ coincides with the ``grounded extension'' \cite{Dung95}). In
fact, $\GE(F)$ can be found via a simple labeling procedure that
repeatedly applies the following two rules to the arguments in $X$ until
each of them is either labeled \IN\ or \OUT:

\newpage
\begin{enumerate}
\item An argument $x$ is labeled \IN\ if all arguments that attack $x$ are
labeled \OUT\ (in particular, if $x$ is not attacked by any
argument).
\item An argument $x$ is labeled \OUT\ if it is attacked by some
  argument that is labeled \IN.
\end{enumerate}
The unique preferred extension $\GE(F)$ is then the set of all
arguments that are labeled~\IN.

\subsection{Value-based systems}

\begin{DEF}
  A \emph{value-based argumentation framework} or \emph{value-based
    system} is a tuple $F=(X,A,V,\eta)$ where $(X,A)$ is an
  argumentation framework, $V$ is a set of \emph{values}, and $\eta$ is
  a mapping $X\rightarrow V$ such that the abstract argumentation system
  $F_v=(\eta^{-1}(v),\SB (x,y)\in A \SM x,y \in \eta^{-1}(v)\SE)$ is acyclic
  for all $v\in V$.

  We call two arguments $x,y\in X$ to be \emph{equivalued} (in $F$) if
  $\eta(x)=\eta(y)$.
\end{DEF}

The requirement for $F_v$ to be acyclic is also known as the
\emph{Multivalued Cycles Assumption}, as it implies that any set of
arguments that form a directed cycle in $F=(X,A)$ will contain at least
two arguments that are not equivalued~\cite{BenchCapon03}.

\begin{DEF}
  An \emph{audience} for a value-based system $F$ is a partial ordering
  $\leq$ on the set $V$ of values of $F$. An audience $\leq$ is
  \emph{specific} if it is a total ordering on $V$.
\end{DEF}
 
For an audience $\leq$ we also define $<$ in the obvious way, i.e.,
$x<y$ if and only if $x\leq y$ and $x \neq y$.

\begin{DEF}
  Given a value-based system $F=(X,A,V,\eta)$ and an audience $\leq$ for
  $F$, we define the abstract argumentation system \emph{induced by
    $\leq$ from $F$} as $F_{\leq}=(X,A_\leq)$ with $A_\leq=\SB (x,y)\in
  A \SM \lnot (\eta(x) < \eta(y)) \SE$.
\end{DEF}

Note that if $\leq$ is a specific audience, then $F_{\leq}=(X,A_\leq)$
is an acyclic system and thus, as discussed above, has a unique
preferred extension $\GE(F_\leq)$.

\begin{EXA}\label{exa:vaf}
  Consider the value-based system $F=(X,A,V,\eta)$ obtained from the
  abstract argumentation framework $F_0$ of Example~\ref{exa:af} by
  adding the set of values $V=\{S,E,T\}$ and the mapping~$\eta$ with
  $\eta(a)=\eta(b)=S$, $\eta(c)=\eta(d)=E$, $\eta(e)=\eta(f)=T$. The
  value-based system $F$ is depicted in Figure~\ref{fig:af} where the
  three ellipses indicate arguments that share the same value.
\end{EXA}

\begin{DEF}
  Let $F=(X,A,V,\eta)$ be a value-based system. We say that an argument
  $x_1\in X$ is \emph{subjectively accepted in $F$} if there exists a
  specific audience $\leq$ such that $x_1$ is in the unique preferred
  extension of $F_\leq$.  Similarly, we say that an argument $x_1\in X$
  is \emph{objectively accepted in $F$} if $x_1$ is contained in the
  unique preferred extension of $F_\leq$ for every specific
  audience~$\leq$.
\end{DEF}

\begin{figure}[tbh]
\centering
  \hspace{5mm}\begin{tikzpicture}[scale=.6]
    \small
    \tikzstyle{every node}=[circle,minimum size=6mm,inner sep=0pt]
    \draw
    (-1.5,0)   node   {$S$:}   
    (-1.5,-2)   node   {$E$:}   
    (-1.5,-4)   node   {$T$:}   

    (0,0)   node[draw] (A) {$a$}   
    (1.5,0) node[draw,fill=lightgray] (B) {$b$} 
    (0,-2)   node[draw,fill=lightgray] (C) {$c$} 
    (1.5,-2) node[draw] (D) {$d$} 
    (1.5,-4) node[draw,fill=lightgray] (F) {$f$} 
    (0,-4)   node[draw,fill=lightgray] (E) {$e$}
    (.75,-5.5) node () {($S>E>T)$} 
    ;
    \draw[thick,->,shorten >=1pt,>=stealth]
    (B) edge (A)
    (C) edge (D)
    (A) edge (D)
    (A) edge[bend right] (E)
    ;
  \end{tikzpicture}\hfill
  \begin{tikzpicture}[scale=.6]
    \small
    \tikzstyle{every node}=[circle,minimum size=6mm,inner sep=0pt]
    \draw
    (0,0)   node[draw] (A) {$a$} 
    (1.5,0) node[draw,fill=lightgray] (B) {$b$} 
    (0,-2)   node[draw] (C) {$c$} 
    (1.5,-2) node[draw,fill=lightgray] (D) {$d$} 
    (1.5,-4) node[draw,fill=lightgray] (F) {$f$} 
    (0,-4)   node[draw,fill=lightgray] (E) {$e$}
    (.75,-5.5) node () {($S>T>E)$} 
    ;
    \draw[thick,->,shorten >=1pt,>=stealth]
    (B) edge (A)
    (C) edge (D)
    (A) edge (D)
    (F) edge (C)
    (A) edge[bend right] (E)
    ;
  \end{tikzpicture}\hfill
  \begin{tikzpicture}[scale=.6]
    \small
    \tikzstyle{every node}=[circle,minimum size=6mm,inner sep=0pt]
    \draw
    (0,0)   node[draw] (A) {$a$} 
    (1.5,0) node[draw,fill=lightgray] (B) {$b$} 
    (0,-2)   node[draw,fill=lightgray] (C) {$c$} 
    (1.5,-2) node[draw] (D) {$d$} 
    (1.5,-4) node[draw,fill=lightgray] (F) {$f$} 
    (0,-4)   node[draw,fill=lightgray] (E) {$e$}
    (.75,-5.5) node () {($E>S>T)$} 
    ;
    \draw[thick,->,shorten >=1pt,>=stealth]
    (B) edge (A)
    (C) edge (D)
    (D) edge (B)
    (A) edge[bend right] (E)
    ;
  \end{tikzpicture}\hfill
  \begin{tikzpicture}[scale=.6]
    \small
    \tikzstyle{every node}=[circle,minimum size=6mm,inner sep=0pt]
    \draw
    (0,0)   node[draw] (A) {$a$} 
    (1.5,0) node[draw,fill=lightgray] (B) {$b$} 
    (0,-2)   node[draw,fill=lightgray] (C) {$c$} 
    (1.5,-2) node[draw] (D) {$d$} 
    (1.5,-4) node[draw,fill=lightgray] (F) {$f$} 
    (0,-4)   node[draw,fill=lightgray] (E) {$e$}
    (.75,-5.5) node () {($E>T>S)$} 
    ;
    \draw[thick,->,shorten >=1pt,>=stealth]
    (B) edge (A)
    (C) edge (D)
    (D) edge (B)
    ;
  \end{tikzpicture}\hfill
  \begin{tikzpicture}[scale=.6]
    \small
    \tikzstyle{every node}=[circle,minimum size=6mm,inner sep=0pt]
    \draw
    (0,0)   node[draw] (A) {$a$} 
    (1.5,0) node[draw,fill=lightgray] (B) {$b$} 
    (0,-2)   node[draw] (C) {$c$} 
    (1.5,-2) node[draw,fill=lightgray] (D) {$d$} 
    (1.5,-4) node[draw,fill=lightgray] (F) {$f$} 
    (0,-4)   node[draw,fill=lightgray] (E) {$e$}
    (.75,-5.5) node () {($T>S>E)$} 
    ;
    \draw[thick,->,shorten >=1pt,>=stealth]
    (B) edge (A)
    (C) edge (D)
    (A) edge (D)
    (F) edge (C)
    ;
  \end{tikzpicture}\hfill
  \begin{tikzpicture}[scale=.6]
    \small
    \tikzstyle{every node}=[circle,minimum size=6mm,inner sep=0pt]
    \draw
    (0,0)   node[draw,fill=lightgray] (A) {$a$} 
    (1.5,0) node[draw] (B) {$b$} 
    (0,-2)   node[draw] (C) {$c$} 
    (1.5,-2) node[draw,fill=lightgray] (D) {$d$} 
    (1.5,-4) node[draw,fill=lightgray] (F) {$f$} 
    (0,-4)   node[draw,fill=lightgray] (E) {$e$}
    (.75,-5.5) node () {($T>E>S)$} 
    ;
    \draw[thick,->,shorten >=1pt,>=stealth]
    (B) edge (A)
    (C) edge (D)
    (D) edge (B)
    (F) edge (C)
    ;
  \end{tikzpicture}\hspace{5mm}
  \caption{Acyclic abstract argumentation system relative to the six
    specific audiences on values $T,S,E$.}\label{fig:six}
\end{figure}
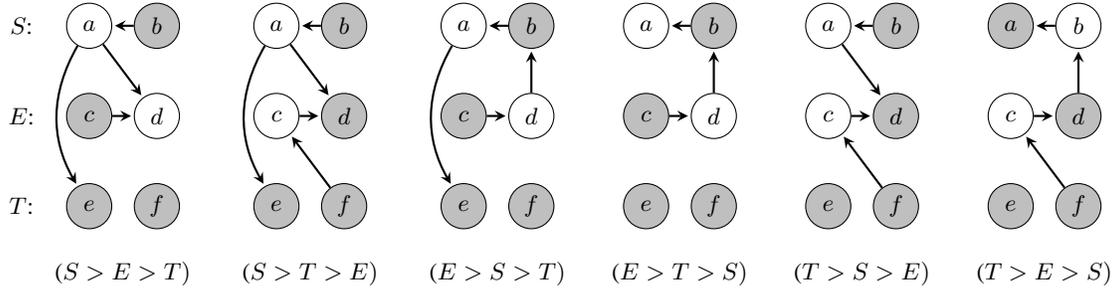

\begin{EXA}
  Consider our running example, the value-based system $F$ given in
  Example~\ref{exa:vaf}.  Suppose $F$ represents the interaction of
  arguments regarding a city development project, and assume the
  arguments $a,b$ are related to sustainability issues ($S$), the
  arguments $c,d$ are related to economics ($E$), and the arguments
  $e,f$ are related to traffic issues ($T$).

  Now, consider the specific audience $\leq$ that gives highest priority
  to sustainability, medium priority to economics, and lowest priority
  to traffic ($S>E>T$). This audience gives rise to the acyclic abstract
  argumentation system $F_\leq$ obtained from $F$ by deleting the attack
  $(d,b)$ (as $\eta(b)=S>E=\eta(d)$, $d$ cannot attack $b$ with respect
  to the audience) and deleting the attack $(f,c)$ (as
  $\eta(c)=E>T=\eta(f)$, $f$~cannot attack $c$ with respect to the
  audience).

  Figure~\ref{fig:six} exhibits the acyclic abstract
  argumentation systems induced by the six possible specific
  audiences. The unique preferred extension for each of the six systems
  is indicated by shaded nodes.  We conclude that all arguments of $F$
  are subjectively accepted, and $e,f$ are the arguments that are
  objectively accepted.
\end{EXA}

\subsection{Computational problems for value-based systems} 
We consider the following decision problems.
\begin{quote}
  \textsc{Subjective Acceptance}
  
  \emph{Instance:} A value-based system $F=(X,A,V,\eta)$ and a query argument $x_1\in
  X$.

  \emph{Question:} Is $x_1$ subjectively accepted in $F$?
\end{quote}

\begin{quote}
  \textsc{Objective Acceptance}

  \emph{Instance:} A value-based system $F=(X,A,V,\eta)$ and a query argument $x_1\in
  X$.

  \emph{Question:} Is $x_1$ objectively accepted in $F$?
\end{quote}
As shown by Dunne and Bench-Capon~\cite{DunneBenchcapon04},
\textsc{Subjective Acceptance} is $\NP$-complete and \textsc{Objective
  Acceptance} is $\coNP$-complete.  Indeed, there are $k!$ possible
specific audiences for a value-based system with $k$ values.  Hence, even
if $k$ is moderately small, say $k=10$, checking all $k!$ induced
abstract argumentation system becomes impractical.  Dunne~\cite{Dunne10}
studied properties of value-based systems that allow to reduce the
number of audiences to consider.

\subsection{Graphical models of value-based systems}\label{subsection:graphs}

In view of the general intractability of \textsc{Subjective Acceptance}
and \textsc{Objective Acceptance}, the main decision problems for value-based systems,
it is natural to ask which restrictions on shape and structure of value-based systems
allow tractability.

A natural approach is to impose structural restrictions in terms of
certain graphical models associated with value-based systems. We present
three graphical models: the \emph{graph structure} (an undirected graph
on the arguments of the value-based system under consideration, edges
represent attacks) the \emph{value graph} (a directed graph on the
values of the value-based system under consideration, edges represent
attacks) and the \emph{extended graph structure} (an undirected graph on
the arguments of the value-based system under consideration, edges
represent attacks and ``equivaluedness''). The concept of value graphs
was recently introduced and studied by Dunne~\cite{Dunne10}. The concept of
\emph{extended graph structures} is our new contribution.

\begin{figure}[tbh]
\centering
  \small
  \hspace{10mm}\begin{tikzpicture}[scale=0.6]
    \tikzstyle{every node}=[circle,minimum size=6mm,inner sep=0pt]
    \draw
    (.75,-5.5) node () {$F$} 
    (0,0)   node[draw] (A) {$a$}  
    (1.5,0) node[draw] (B) {$b$} 
    (0,-2)   node[draw] (C) {$c$} 
    (1.5,-2) node[draw] (D) {$d$} 
    (1.5,-4) node[draw] (F) {$f$} 
    (0,-4)   node[draw] (E) {$e$}
    ;
    \draw[thick,->,shorten >=1pt,>=stealth]

    (B) edge (A)
    (C) edge (D)
    (D) edge (B)
    (A) edge (D)
    (F) edge (C)
    (A) edge[bend right] (E)
    ;
    \draw  (0.75,0) ellipse (1.95 and 0.85);
    \draw  (0.75,-2) ellipse (1.95 and 0.85);
    \draw  (0.75,-4) ellipse (1.95 and 0.85);
    \draw  (-1.6,0) node () {$S$:} ;
    \draw  (-1.6,-2) node () {$E$:} ;
    \draw  (-1.6,-4) node () {$T$:} ;
  \end{tikzpicture}\hfill
  \begin{tikzpicture}[scale=0.6]
    \tikzstyle{every node}=[circle,minimum size=6mm,inner sep=0pt]
    \draw
    (.75,-5.5) node () {$G_F$} 
    (0,0)   node[draw] (A) {$a$}  
    (1.5,0) node[draw] (B) {$b$} 
    (0,-2)   node[draw] (C) {$c$} 
    (1.5,-2) node[draw] (D) {$d$} 
    (1.5,-4) node[draw] (F) {$f$} 
    (0,-4)   node[draw] (E) {$e$}

    ;
    \draw[thick]

    (B) edge (A)
    (C) edge (D)
    (D) edge (B)
    (A) edge (D)
    (F) edge (C)
    (A) edge[bend right] (E)
    ;
  \end{tikzpicture}\hfill
  \begin{tikzpicture}[scale=0.6]
    \tikzstyle{every node}=[circle,minimum size=6mm,inner sep=0pt]
    \draw
    (0,0)   node[draw] (S) {$S$}  
    (0,-2)   node[draw] (T) {$E$} 
    (0,-4) node[draw] (E) {$T$} 
    (0,-5.5) node () {$G_F^\val$} 
    ;
    \draw[thick,->,shorten >=1pt,>=stealth]
    (S) edge[bend right] (T)
    (T) edge[bend right] (S)
    (E) edge (T)
    (S) edge[bend right] (T)
    (S) edge[bend right=45mm] (E)
    ;
  \end{tikzpicture}\hfill
  \begin{tikzpicture}[scale=0.6]
    \tikzstyle{every node}=[circle,minimum size=6mm,inner sep=0pt]
    \draw
    (.75,-5.5) node () {$G_F^\ext$} 
    (0,0)   node[draw] (A) {$a$}  
    (1.5,0) node[draw] (B) {$b$} 
    (0,-2)   node[draw] (C) {$c$} 
    (1.5,-2) node[draw] (D) {$d$} 
    (1.5,-4) node[draw] (F) {$f$} 
    (0,-4)   node[draw] (E) {$e$}
 
    ;
    \draw[thick]

    (B) edge (A)
    (C) edge (D)
    (D) edge (B)
    (A) edge (D)
    (F) edge (C)
    (A) edge[bend right] (E)
    (E) edge (F)
    ;
  \end{tikzpicture}\hspace{20mm}
\caption{A value-based system $F$ with its graph
  structure $G_F$, value graph $G_F^\val$, and extended graph structure~$G_F^\ext$.}\label{fig:graphs}
\end{figure}
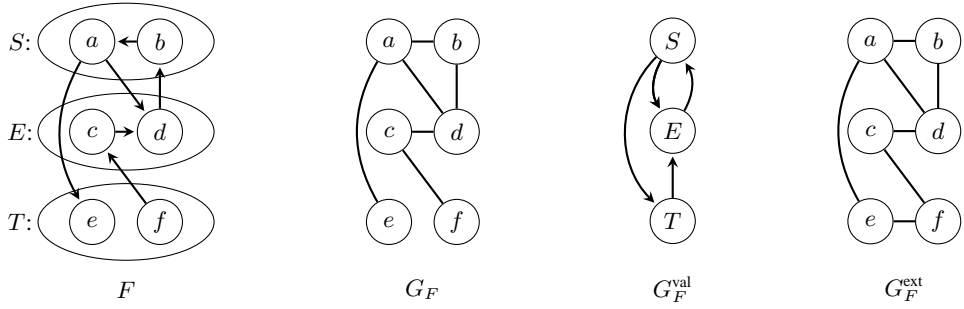

\begin{DEF}\label{def:graphs}
  Let $F=(X,A,V,\eta)$ be a value-based system. 

  The \emph{graph structure} of $F$ is the (undirected) graph $G_F=(X,E)$
  whose vertices are the arguments of $F$ and where two arguments $x,y$
  are joined by an edge (in symbols $xy\in E$) if and only if $X$
  contains the attack $(x,y)$ or the attack $(y,x)$.

  The \emph{value graph} of $F$ is the directed graph $G_F^\val=(V,E)$
  whose vertices are the values of~$F$ and where two values $u,v$ are
  joined by a directed edge from $u$ to $v$ (in symbols $(u,v)\in E$) if
  and only if there exist some argument $x\in X$ with $\eta(x)=u$, some
  argument $y\in X$ with $\eta(y)=v$, and $(x,y)\in A$.

  The \emph{extended graph structure} of $F$ is the (undirected) graph
  $G_F^\ext=(X,E)$ whose vertices are the arguments of $F$ and where two
  arguments $x,y$ are joined by an edge if and only if $(x,y)\in A$ or
  $\eta(x)=\eta(y)$.
\end{DEF}
Figure~\ref{fig:graphs} shows the value-based system of Example~\ref{exa:vaf} and the
three associated graphical models.

\begin{DEF}
  A value-based system $F=(X,A,V,\eta)$ is called \emph{bipartite} if its graph
  structure is a bipartite graph, i.e., if $X$ can be partitioned into two
  conflict-free sets.
\end{DEF}

\section{Value-width and attack-width}\label{sec:valuewidth}
\noindent Dunne~\cite{Dunne07} suggested to consider restrictions on the
number of arguments that share the same value, and the number of attacks
between equivalued arguments. We state these restrictions in terms of
the following notions.

\begin{DEF}
  Let $F=(X,A,V,\eta)$ be a value-based system. The \emph{value-width} of $F$ is the
  largest number of arguments that share the same value, i.e.,
  $\max_{v\in V} \Card{\eta^{-1}(v)}$.  The \emph{attack-width} of $F$
  is the cardinality of the set $\SB (x,y)\in A \SM \eta(x)=\eta(y)
  \SE$.
\end{DEF}

For instance, the value-based system of Example~\ref{exa:vaf} has
value-width 2 and attack-width~2.

\emph{Value-based systems of value-width 1} are not very interesting:
Every argument $x$ in such a value-based system is subjectively accepted
($x$ is accepted with respect to any specific audience where $\eta(x)$ is
largest), and objectively accepted if and only if $x$ is not attacked by
any argument $y$ (if $y$ attacks $x$ then $x$ is not accepted with
respect to any specific audience where $\eta(y)$ is largest). Thus, for
value-based systems of value-width 1 the problems \textsc{Subjective}
and \textsc{Objective Acceptance} are trivial, and the expressive power
of such value-based systems is very limited.

On the other hand, \emph{value-based systems of value-width 3} are
already too expressive to allow a tractable acceptance decision:
Dunne~\cite{Dunne07} showed that the problems \textsc{Subjective} and
\textsc{Objective Acceptance} are intractable ($\NP$-complete and
$\coNP$-complete, respectively) for value-based systems of
value-width~3, even if their graph structure is a tree.

This leaves the intermediate class of \emph{value-based systems of
  value-width 2} as an interesting candidate for a tractable class.  In
fact, Dunne~\cite{Dunne07} conjectured that both acceptance problems are
polynomial-time decidable for value-based systems of value-width 2. He
also conjectured that the problems are polynomial for value-based
systems with an attack-width that is bounded by a constant.  We disprove
both conjectures and show that the problems remain intractable for
value-based systems of value-width 2 and (simultaneously) of
attack-width~1.

On the positive side, we show that under the additional assumption that
the value-based system is bipartite (that entails value-based systems
whose graph structures are trees) both acceptance problems can be
decided in polynomial time for value-based systems of value-width~2.

\begin{THE}\label{size2hard}
  \normalfont{(A)} \textsc{Subjective Acceptance} remains $\NP$-hard for
  value-based systems of value-width~2 and attack-width~1.
  \normalfont{(B)} \textsc{Objective Acceptance} remains $\coNP$-hard for
  value-based systems of value-width~2 and attack-width~1.
\end{THE}

\enlargethispage{5mm}

\begin{THE}\label{bipeasy}
  \normalfont{(A)} \textsc{Subjective Acceptance} can be decided in polynomial
  time for bipartite value-based systems  of value-width~2.
  \normalfont{(B)} \textsc{Objective Acceptance} can be decided in
  polynomial time for bipartite value-based systems of value-width~2.
\end{THE}
In the remainder of this section we will demonstrate the two theorems.

\subsection{Certifying paths}\label{sec:cert}

The key to the proofs of Theorems~\ref{size2hard} and \ref{bipeasy} is
the notion of a ``certifying path'' which defines a certain path-like
substructure within a value-based system. We show that in value-based
systems of value-width 2, the problems of \textsc{Subjective} and
\textsc{Objective Acceptance} can be expressed in terms of certifying
paths. We then show that in general finding a certifying path in a
value-based system of value-width 2 is $\NP$-hard (3SAT can be expressed
in terms of certifying paths) but is easy if the system is bipartite.

\begin{DEF}\label{def:cp}
   Let $F=(X,A,V,\eta)$ be a value-based system of value-width~$2$.  We call an
   odd-length sequence $C=(x_1,z_1,\dots,x_{k},z_{k},t)$, $k\geq 0$, of
   distinct arguments a \emph{certifying path for $x_1\in X$ in~$F$} if
   it satisfies the following conditions:
   \begin{enumerate}
   \item[C1] For every $1 \leq i \leq k$ it holds that
     $\eta(z_i)=\eta(x_i)$.
   \item[C2] For every $1 \leq i \leq k$ there exists a $1 \leq j \leq
     i$ such that $z_i$ attacks $x_j$.
   \item[C3] For every $2 \leq i \leq k$ it holds that $x_i$ attacks
     $z_{i-1}$ but $x_i$ does not attack any argument in
     $\{z_i,x_1,\dots,x_{i-1}\}$.
   \item[C4] Argument $t$ attacks $z_{k}$ but it does not attack any argument in
     $\{x_1,\dots,x_k\}$.
   \item[C5] If there exists an argument $z\in X\setminus \{t\}$ with
     $\eta(z)=\eta(t)$ then either $t$ attacks $z$ or $z$ does not
     attack any argument in $\{x_1,\dots,x_k,t\}$.
   \end{enumerate}
\end{DEF}

\begin{LEM}\label{certpathsa}
  Let $F=(X,A,V,\eta)$ be a value-based system of value-width~2 and $x_1 \in X$. Then $x_1$ is
  subjectively accepted in $F$ if and only if there exists a certifying path
  for $x_1$ in $F$.
\end{LEM}

The rather technical proof of this lemma is given in the
appendix. We discuss the intuition behind the concept of certifying paths by
means of an example.

\begin{EXA}
  Consider the value-based system $F$ of Example~\ref{exa:vaf}. We want to check
  whether argument $a$ is subjectively accepted, i.e., to identify a
  specific audience $\leq$ such that $a$ is in the unique preferred
  extension $\GE(F_\leq)$ of~$F_\leq$. Since $a$ is attacked by $b$ and
  we cannot eliminate this attack ($a$ and $b$ are equivalued), we need
  to defend~$a$ by attacking~$b$. The only possibility for that is to
  attack $b$ by $d$. Hence we need to put $S<E$ in our
  audience. However, since $d$ is attacked by the equivalued argument
  $c$, we need to defend it by attacking $c$ by $f$, hence we need to
  put $S<E<T$. Since $f$ is not attacked by any other argument we can
  stop. Via this process we have produced a certifying path
  $C_a=(a,b,d,c,f)$, and we can check that $C_a$ indeed satisfies
  Definition~\ref{def:cp}. For the other subjectively accepted arguments
  of $F$ we have the certifying paths $C_b=(b)$, $C_c=(c)$,
  $C_d=(d,c,f)$, $C_e=(e)$ and $C_f=(f)$.
\end{EXA}

In order to use the concept of certifying paths for objective
acceptance, we need the following definition.

\begin{DEF}
  Let $F=(X,A,V,\eta)$ be a value-based system and $v \in V$ a value. We
  denote by $F-v$ the value-based system obtained from~$F$ by deleting
  all arguments with value $v$ and all attacks involving these
  arguments.
\end{DEF}

\begin{LEM}\label{certpathoa}
  Let $F=(X,A,V,\eta)$ be a value-based system of value-width~2 and $x_1
  \in X$. Then $x_1$ is objectively accepted in $F$ if and only if for
  every argument $p\in X$ that attacks $x_1$ it holds that $\eta(p)\neq
  \eta(x_1)$ and $p$ is \emph{not} subjectively accepted in
  $F-\eta(x_1)$.
\end{LEM}
Again, the technical proof is moved to the appendix. 

\begin{EXA}
  In our example, consider the argument $e$. We want to check whether
  $e$ is objectively accepted. Since~$e$ is only attacked by $a$, and
  since $\eta(a)\neq \eta(e)$, it remains to check whether $a$ is not
  subjectively accepted in $F-\eta(e)$. In fact, $F-\eta(e)$ contains no
  certifying path for $a$. Hence $e$ is objectively accepted~in $F$.
\end{EXA}

\subsection{Hardness for value-based systems of value-width 2}\label{subsection:hard}

\begin{figure}[tbh]
\centering\small
   \begin{tikzpicture}[xscale=0.9,yscale=1.5]
    \small
    \tikzstyle{every node}=[circle,minimum size=6mm,inner sep=0pt]
    \draw

    (0,0) ellipse (1.1 and 0.45)
    (0,-1) ellipse (1.1 and 0.45)
    (0,-2) ellipse (1.1 and 0.45)
    (0,-3) ellipse (1.1 and 0.45)
    (0,-4) ellipse (1.1 and 0.45)
    (0,-5) ellipse (1.1 and 0.45)
    (0,-6) ellipse (1.1 and 0.45)

    (-2.5,-1) ellipse (1.1 and 0.45)
    (2.5,-1) ellipse (1.1 and 0.45)

    (-2.5,-3) ellipse (1.1 and 0.45)
    (2.5,-3) ellipse (1.1 and 0.45)

    (-2.5,-5) ellipse (1.1 and 0.45)
    (2.5,-5) ellipse (1.1 and 0.45)

    (-0.5,0)   node[draw] (z1) {$z_1$} 
    (.5,0)   node[draw] (x1) {$x_1$} 
    (.5,-2)   node[draw] (x2) {$x_2$} 
    (-.5,-2)   node[draw] (z2) {$z_2$} 
    (.5,-4)   node[draw] (x3) {$x_3$} 
    (-.5,-4)   node[draw] (z3) {$z_3$} 
    (0,-6)   node[draw] (t) {$t$}

    (-2.0,-1)   node[draw] (z11) {$z_1^1$} 
    (-3.0,-1)   node[draw] (x11) {$x_1^1$} 
    (.5,-1)   node[draw] (z12) {$z_1^2$} 
    (-.5,-1)   node[draw] (x12) {$x_1^2$} 
    (3.0,-1)   node[draw] (z13) {$z_1^3$} 
    (2.0,-1)   node[draw] (x13) {$x_1^3$}

    (-2.0,-3)   node[draw] (z21) {$z_2^1$} 
    (-3.0,-3)   node[draw] (x21) {$x_2^1$} 
    (.5,-3)   node[draw] (z22) {$z_2^2$} 
    (-.5,-3)   node[draw] (x22) {$x_2^2$} 
    (3.0,-3)   node[draw] (z23) {$z_2^3$} 
    (2.0,-3)   node[draw] (x23) {$x_2^3$}

    (-2.0,-5)   node[draw] (z31) {$z_3^1$} 
    (-3.0,-5)   node[draw] (x31) {$x_3^1$} 
    (.5,-5)   node[draw] (z32) {$z_3^2$} 
    (-.5,-5)   node[draw] (x32) {$x_3^2$} 
    (3.0,-5)   node[draw] (z33) {$z_3^3$} 
    (2.0,-5)   node[draw] (x33) {$x_3^3$}

;

     \draw[->,shorten >=.5pt,>=stealth]

     (z1) edge (x1)

     (x11) edge (z1)
     (z11) edge (x1)
     (x12) edge (z1)
     (z12) edge (x1)
     (x13) edge (z1)
     (z13) edge (x1)
     (z2) edge (x11) 
     (x2) edge (z11) 
     (z2) edge (x12) 
     (x2) edge (z12) 
     (z2) edge (x13) 
     (x2) edge (z13) 

     (x21) edge (z2)
     (z21) edge (x2)
     (x22) edge (z2)
     (z22) edge (x2)
     (x23) edge (z2)
     (z23) edge (x2)
     (z3) edge (x21) 
     (x3) edge (z21) 
     (z3) edge (x22) 
     (x3) edge (z22) 
     (z3) edge (x23) 
     (x3) edge (z23) 

     (x31) edge (z3)
     (z31) edge (x3)
     (x32) edge (z3)
     (z32) edge (x3)
     (x33) edge (z3)
     (z33) edge (x3)
     (t) edge (z31) 
     (t) edge (z32) 
     (t) edge (z33) 

     ;

      \draw[->,thick]
      (x21) edge[bend left] (x11)
      (x23) edge[bend left] (x13)
      (x31) edge[bend left] (x21)
      (x32) edge[bend left=60] (x22)
      (x32) edge[bend right=27] (x12)
      (x33) edge[bend right=25] (x13)
;

  \end{tikzpicture}
  \caption{The value-based system $F$ in the proof of
    Theorem~\ref{size2hard} for the 3CNF Formula $(x_1 \lor x_2 \lor
    x_3) \land ( \lnot x_1 \lor x_2 \lor \lnot x_3 ) \land ( x_1 \lor
    \lnot x_2 \lor \lnot x_3 )$.}
  \label{fig:hardness}
\end{figure}
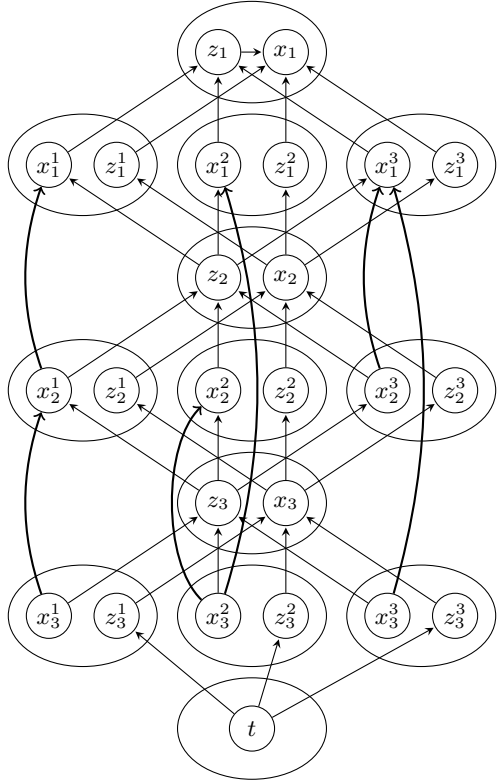

This subsection is devoted to prove Theorem~\ref{size2hard}.  We devise
a polynomial reduction from {\sc 3SAT}.  Let $\Phi$ be a 3CNF formula
with clauses $C_1,\dots,C_m$ and $C_j=x_{j,1} \vee x_{j,2} \vee
x_{j,3}$ for $1 \leq j \leq m$.  In the following we construct a value-based system
$F=(X,A,V,\eta)$ of value-width~2 and attack-width~1 such that the query
argument $x_1 \in X$ is subjectively accepted in $F$ if and only if
$\Phi$ is satisfiable. See Figure~\ref{fig:hardness} for an example.

The set $X$ contains the following arguments: 
\begin{enumerate}
\item a pair of arguments $x_j,z_j$  for $1\leq j
  \leq m$;
\item a pair of arguments $x_j^i,z_j^i$ for $1\leq j \leq m$ and $1
  \leq i \leq 3$;
\item an argument $t$.
\end{enumerate}

The set $A$ contains the following attacks:
\begin{enumerate}
\item  $(z_1,x_1)$;
\item  $(x_j^i,z_j)$ and $(z_j^i,x_j)$ for  $1 \leq j
  \leq m$ and $1 \leq i \leq 3$;
\item  $(x_{j+1},z_j^i)$ and $(z_{j+1},x_j^i)$ for $1 \leq j
  \leq m-1$ and $1 \leq i \leq 3$;
\item  $(t,z_m^i)$ for  $1 \leq i \leq 3$;
\item  $(x_j^i,x_{j'}^{i'})$ for $1 \leq j' < j \leq m$ and
  $1 \leq i,i' \leq 3$ whenever $x_{j,i}$ and $x_{j',i'}$ are
  complementary literals.
\end{enumerate}
The set $V$ contains one value for each $x,z$ pair, and one value for
argument $t$, i.e., $\Card{V}=4m+1$.  Consequently, the mapping $\eta$
is defined such that $\eta(x_j)=\eta(z_j)=v_j$,
$\eta(x_j^i)=\eta(z_j^i)=v_j^i$ for $1 \leq j \leq m$, $1 \leq i \leq
3$, and $\eta(t)=v_t$.  Evidently $F$ has attack-width~1 and
value-width~2, and it is clear that $F$ can be constructed from $\Phi$
in polynomial time.

We establish part (A) of Theorem~\ref{size2hard}
by showing the following claim.
\begin{CLM}\label{claim:subj}
  $\Phi$ is satisfiable if and only if $x_1$ is subjectively accepted in
  $F$.
\end{CLM}
\begin{proof}
  First we note that every certifying path for $x_1$ in $F$ must have
  the form 
  $(x_1$, $z_1$, 
  $x_1^{i_1}$, $z_1^{i_1}$,
  $x_2$, $z_2$,  
  $x_2^{i_2}$, $z_2^{i_2}$, 
  $x_3$, $z_3,   \dots,
  x_m$, $z_m$, 
  $x_m^{i_m}$, $z_m^{i_m}$, $t)$ where  $i_j \in \{1,2,3\}$ for every
  $1 \leq j \leq m$ and for every pair $1 \leq j < j' \leq m$ there is
  no attack $(x_{j'}^{i_{j'}},x_j^{i_j})\in A$. Hence
  there exists a certifying path for $x_1$ in $F$ if and only if there
  exists a set $L$ of literals that corresponds to a satisfying truth
  assignment of~$\Phi$ (i.e., $L$ contains a literal of each clause of
  $\Phi$ but does not contain a complementary pair of literals).
\end{proof}

In order
to show part (B) of  Theorem \ref{size2hard},
let $F$ be the value-based system as constructed above and define 
$F'=(X',A',V',\eta')$ to be the value-based system with
\begin{enumerate}
\item  $X':=X \cup \{x_0\}$,
\item  $A':=A \cup \{(x_1,x_0)\}$, 
\item $V':=V \cup \{v_0\}$, 
\item $\eta'(x_0)=v_0$ and  $\eta'(x)=\eta(x)$ for every $x \in X$.
\end{enumerate}

Part (B) of Theorem~\ref{size2hard} follows from the following claim
which follows from Claim~\ref{claim:subj} and Lemma~\ref{certpathoa}.
\begin{CLM}\label{claim:obj}
  $\Phi$ is satisfiable if and only if $x_0$ is not objectively accepted in
  $F'$.
\end{CLM}

By a slight modification of the above reduction we can also show the
following, answering a research question recently posed by
Dunne~\cite{Dunne10}. The detailed argument is given in the appendix.

\begin{COR}\label{cor:bip}
  \textsc{Subjective} and \textsc{Objective Acceptance} remain
  $\NP$-hard and $\coNP$-hard, respectively, for value-based systems
  whose value graphs are bipartite.
\end{COR}

\subsection{Certifying paths and dialogues}

Bench-Capon, Doutre, and Dunne~\cite{BenchcaponDoutreDunne07} developed
a general \emph{dialogue framework} that allows to describe the
acceptance of arguments in a value-based system in terms of a game,
played by two players, the proponent and the opponent. The proponent
tries to prove that a certain argument (or a set of arguments) is
accepted, the opponent tries to circumvent the proof. An argument is
subjectively accepted if the proponent has a winning strategy, that is,
she is able to prove the acceptance regardless of her opponent's moves.

In the following we outline a simplified version of the dialogue framework
that applies to value-based systems of value-width~2. We
will see that certifying paths correspond to winning strategies for the
proponent.

Let $F=(X,A,V,\eta)$ be a value-based system of value-width $2$. We have
two players, the proponent and the opponent, who make moves in turn, at
each move asserting a new argument. This produces a sequence
$(x_1,y_1,x_2,y_2\dots)$ of arguments and a set of audiences $\leq$ with
$\eta(x_1)=\eta(y_2) < \eta(x_2)=\eta(y_2) < \dots $.  The proponent has
the first move, where she asserts the query argument $x_1$ whose
subjective acceptance is under consideration.  After each move of the
proponent, asserting argument $x_i$, the opponent asserts a new argument
$y_i\in X\setminus\{x_1,y_1,\dots,x_{i-1},y_{i-1},x_i\}$ which has the
same value as $x_i$ but is not attacked by $x_i$, and attacks some
argument asserted by the proponent.  If no such argument $y_i$ exists,
the proponent has won the game. After each move of the opponent
asserting an argument $y_i$, it is again the proponent's turn to assert
a new argument $x_{i+1}\in X\setminus \{x_1,y_1,\dots,x_i,y_i\}$. This
argument $x_{i+1}$ must attack the opponent's last argument $y_i$, but
must not attack any argument asserted by the proponent. If no such
argument $x_{i+1}$ exists, the proponent has lost the game. Because the
value-width of $F$ is assumed to be $2$, the opponent has at most one
choice for each move. Therefore, the proponent's wining strategy does
not need to consider several possibilities for the opponent's counter
move. Hence, a winning strategy is not a tree but just a path and can be
identified with a sequence $(x_1,y_1,\dots,x_{n-1},y_{n-1},x_n)$ that
corresponds to a play won by the proponent. It is easy to verify that
such a sequence is exactly a certifying path.

\begin{EXA}\label{exa:game}
  Consider again the value-based system $F$ of
  Example~\ref{exa:vaf}. The proponent wants to prove that argument $a$
  is subjectively accepted in~$F$ and asserts $a$ with her first
  move. Now, it is the opponent's turn. He has no other choice but to
  assert $b$ (the only argument different from $a$ with the same value
  as~$a$). Now, it is again the proponent's turn. She must assert an
  argument $x \notin \{a,b\}$ that attacks $b$ but does not attack
  $a$. Argument $d$ satisfies this property (it happens that this is the
  only choice). Next, the opponent asserts $c$, and the proponent
  asserts $f$, and it is again the opponent's turn. The only argument
  with the same value as $f$ is argument $e$, but $e$ does not attack
  any of the arguments in $\{a,d,f\}$. Hence, the proponent wins. The
  sequence of arguments $(a,b,d,c,f)$ produced by this play is indeed a
  certifying path for $a$ in~$F$. Hence $a$ is subjectively accepted.
\end{EXA}

\subsection{Polynomial-time algorithm for bipartite value-based systems
  of value-width 2}

In this subsection we prove Theorem \ref{bipeasy}. Throughout this
section, we assume that we are given a bipartite value-based system $F=(X,A,V,\eta)$
together with a query argument $x_1$. Furthermore, let $X_{\text{even}}$
and $X_{\text{odd}}$ be the subsets of $X$ containing all arguments $x$
such that the length of a shortest directed path in $F$ from $x$ to
$x_1$ is even and odd, respectively.
\begin{LEM}\label{parity}
  Let $C=(x_1,z_1,\dots,x_k,z_k,t)$ be a certifying path for $x_1$ in
  $F$.  Then $\SB x_i \SM 1 \leq i \leq k \SE \cup \{t\} \subseteq
  X_{\mathrm{even}}$ and $\SB z_i \SM 1 \leq i \leq k \SE \subseteq
  X_{\mathrm{odd}}$.
\end{LEM}
\begin{proof} 
The claim follows easily via induction on $k$ by using the properties of
a certifying path and the fact that $F$ is bipartite.
\end{proof}
Based on the observation of Lemma \ref{parity}, we construct an
auxiliary directed graph $H_{F}:=(V,E)$ as follows. The vertex set of
$H_F$ is the set $V$ of values of $F$. There is a directed edge from $u$
to $v$ if and only if there is an argument $x \in X_{\text{even}}$ with
$\eta(x)=u$ and an argument $z$ with $\eta(z)=v$ such that $(x,z) \in
A$.  Note that $z \in X_{\text{odd}}$ since $F$ is bipartite.
 
\begin{LEM}\label{path1}
  If $C=(x_1,z_1,\dots,x_k,z_k,t)$ is a certifying path for $x_1$ in $F$,
  then $(\eta(t),\eta(x_k), \dots,\eta(x_1))$ is a directed 
  path from $\eta(t)$ to $\eta(x_1)$ in $H_F$.
\end{LEM}
\begin{proof}
  By the definition of a certifying path, we have $(t,z_k) \in A$ and
  for every $2 \leq i \leq k$ it holds that $(x_i,z_{i-1})\in A$. Lemma
  \ref{parity} implies that for $t$ and $x_i$ are contained in
  $X_{\text{even}}$ for every $1 \leq i \leq k$, and hence
  $(\eta(t),\eta(x_k)),(\eta(x_i),\eta(x_{i-1})) \in E$ for every $1 < i
  \leq k$.
\end{proof}

Lemma \ref{path1} tells us that each certifying path in $F$ gives rise to a
directed path in $H_F$.

\begin{figure}[tbh]
\begin{center}
  \hspace{5mm}\begin{tikzpicture}[xscale=.8,yscale=.6]
    \small
    \tikzstyle{every node}=[circle,minimum size=6mm,inner sep=0pt]
    \draw
    (1.25,-10) node () {$F$} 
    (0,0)   node[draw] (x1) {$x_1$} 
    (2.5,0) node[draw] (z1) {$z_1$} 
    (0,-2)   node[draw] (x2) {$x_2$} 
    (2.5,-2) node[draw] (z2) {$z_2$} 
    (0,-4)   node[draw] (x3) {$x_3$} 
    (2.5,-4)   node[draw] (z3) {$z_3$}
    (0,-6) node[draw] (x4) {$x_4$} 
    (2.5,-6)   node[draw] (z4) {$z_4$}
    (0,-8) node[draw] (x5) {$x_5$} 
    (2.5,-8)   node[draw] (z5) {$z_5$}
    ;
   \draw  (1.25,0) ellipse (1.95 and 0.90)  +(-2.3,0) node () {$v_1$:};
   \draw  (1.25,-2) ellipse (1.95 and 0.90) +(-2.3,0) node () {$v_2$:};
   \draw  (1.25,-4) ellipse (1.95 and 0.90) +(-2.3,0) node () {$v_3$:};
   \draw  (1.25,-6) ellipse (1.95 and 0.90) +(-2.3,0) node () {$v_4$:};
   \draw  (1.25,-8) ellipse (1.95 and 0.90) +(-2.3,0) node () {$v_5$:};

    \draw[thick,->,shorten >=1pt,>=stealth]
    (z1) edge (x1)
    (x2) edge (z1)
    (x2) edge (z5)
    (x3) edge (z2)
    (x5) edge (z4)
    (x5) edge (z3)
    (x4) edge (z2)
    (z3) edge (x3)
    (z2) edge (x1)
    (z4) edge (x2)
     ;
    \draw
    (6,-10) node () {$H_F$} 
    (6,0)   node[draw] (v1) {$v_1$} 
    (6,-2)   node[draw] (v2) {$v_2$} 
    (6,-4)   node[draw] (v3) {$v_3$} 
    (6,-6) node[draw] (v4) {$v_4$} 
    (6,-8) node[draw] (v5) {$v_5$} 
    ;
    \draw[thick,->,shorten >=1pt,>=stealth]
    (v5) edge [bend left] (v3) 
    (v3) edge (v2) 
    (v2) edge  (v1)
    (v5) edge (v4)
    (v4) edge [bend left] (v2)
    (v2) edge [bend left] (v5)
    ;
    \draw
    (8,-10) node () {$H_F^{-v_1}$} 
    (8,-2)   node[draw] (v2) {$v_2$} 
    (8,-4)   node[draw] (v3) {$v_3$} 
    (8,-6) node[draw] (v4) {$v_4$} 
    (8,-8) node[draw] (v5) {$v_5$} 
    ;
    \draw[thick,->,shorten >=1pt,>=stealth]
    (v5) edge [bend left] (v3) 
    (v3) edge (v2) 
       (v4) edge [bend left] (v2)
    (v5) edge (v4)
    (v2) edge [bend left] (v5)
    ;
    \draw
    (10,-10) node () {$H_F^{-v_2}$} 
    (10,-2)   node[draw] (v2) {$v_2$} 
    (10,-4)   node[draw] (v3) {$v_3$} 
    (10,-6) node[draw] (v4) {$v_4$} 
    (10,-8) node[draw] (v5) {$v_5$} 
    ;
    \draw[thick,->,shorten >=1pt,>=stealth]
    (v5) edge [bend left] (v3) 
    (v3) edge (v2) 
    (v5) edge (v4)
    (v4) edge [bend left] (v2)
    (v2) edge [bend left] (v5)
    ;
    \draw
    (12,-10) node () {$H_F^{-v_3}$} 
    (12,0)   node[draw] (v1) {$v_1$} 
    (12,-2)   node[draw] (v2) {$v_2$} 
    (12,-6) node[draw] (v4) {$v_4$} 
    (12,-8) node[draw] (v5) {$v_5$} 
    ;
    \draw[thick,->,shorten >=1pt,>=stealth]
    (v2) edge  (v1)
    (v5) edge (v4)
    (v4) edge [bend left] (v2)
    (v2) edge [bend left] (v5)
    ;
    \draw
    (14,-10) node () {$H_F^{-v_4}$} 
    (14,0)   node[draw] (v1) {$v_1$} 
    (14,-4)   node[draw] (v3) {$v_3$} 
    (14,-6) node[draw] (v4) {$v_4$} 
    (14,-8) node[draw] (v5) {$v_5$} 
    ;
    \draw[thick,->,shorten >=1pt,>=stealth]
    (v5) edge [bend left] (v3) 
    (v5) edge (v4)
    ;
    \draw
    (16,-10) node () {$H_F^{-v_5}$} 
    (16,0)   node[draw] (v1) {$v_1$} 
    (16,-2)   node[draw] (v2) {$v_2$} 
    (16,-4)   node[draw] (v3) {$v_3$} 
    (16,-6) node[draw] (v4) {$v_4$} 
    (16,-8) node[draw] (v5) {$v_5$} 
    ;
    \draw[thick,->,shorten >=1pt,>=stealth]
    (v5) edge [bend left] (v3) 
    (v3) edge (v2) 
    (v2) edge  (v1)
    (v5) edge (v4)
    (v4) edge [bend left] (v2)
    (v2) edge [bend left] (v5)
    ;
  \end{tikzpicture}\hspace{5mm}
\end{center}
\caption{A bipartite value-based system $F$, the associated axillary
  graph $H_F$, and various subgraphs obtained by deleting a
  value.}\label{fig:aux}
\end{figure}
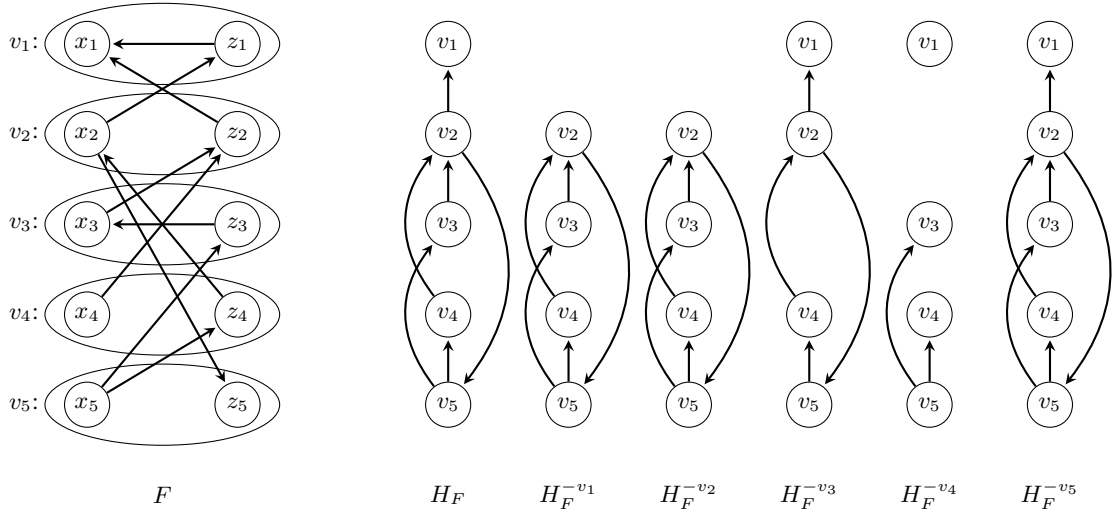

\begin{EXA}\label{exa:aux}
  Figure~\ref{fig:aux} shows a bipartite value-based system $F$ and the
  associated auxiliary graph $H_F$. The query argument is $x_1$. Hence
  $X_{\text{even}}=\{x_1,\dots,x_5\}$ and
  $X_{\text{odd}}=\{z_1,\dots,z_5\}$. The query argument $x_1$ is
  subjectively accepted in $F$ as $C=(x_1,z_1,x_2,z_2,x_4,z_4,x_5)$ is a
  certifying path for $x_1$ in $F$. Indeed, $C$ gives rise to the
  directed path $v_5,v_4,v_2,v_1$ (i.e.,
  $\eta(x_5),\eta(x_4),\eta(x_2),\eta(x_1)$) in $H_F$, as
  promised by Lemma \ref{path1}.
\end{EXA}

It would be desirable if we could find certifying paths by searching for
directed paths in $H_F$. However, not every directed path in $H_F$ gives
rise to a certifying path in $F$.  To overcome this obstacle, we
consider for each value $v\in V$ the subgraph $H_F^{-v}$ of $H_{F}$
which is obtained as follows:

If there is an argument $z\in X_{\text{odd}}\cap \eta^{-1}(v)$ that is
not attacked by some equivalued argument, then for every argument $y\in
X_{\text{even}}$ that is attacked by $z$ we remove the vertex $\eta(y)$
from $H_{F}$.

Figure~\ref{fig:aux} shows the graphs $H_F^{-v}$ for the value-based system $F$ of
Example~\ref{exa:aux}.

\begin{LEM}\label{path2}
  Consider an odd-length sequence $C=(x_1,z_1,\dots,x_k,z_k,t)$ of distinct
  arguments of a bipartite value-based system $F$ of value width $2$.  Then $C$ is a
  certifying path for $x_1$ in $F$ if and only if the following conditions
  hold:
\begin{enumerate}
\item[(1)] $\eta(x_i)=\eta(z_i)$ for $1\leq i \leq k$.
\item[(2)] $(\eta(t), \eta(x_k),\ldots ,\eta(x_1))$ is a directed
  path from $\eta(t)$ to $\eta(x_1)$ in $H_F^{-\eta(t)}$.
\item[(3)] None of the sub-sequences $\eta(x_i),\dots ,\eta(x_1)$ is a
  directed path from $\eta(x_i)$ to $\eta(x_1)$ in $H_F^{-\eta(x_i)}$ for \hbox{$1
  \leq i \leq k$}.
\end{enumerate}
\end{LEM}
\begin{proof}
Assume $C=(x_1,z_1,\dots,x_k,z_k,t)$ is a certifying path for $x_1$ in $F$.
Property (1) follows from condition C1 of a certifying path, property (2)
follows from condition C5 and Lemma~\ref{path1}. Property (3) follows from
conditions C2 and~C3.

To see the reverse assume that $C$ satisfies properties (1)--(3).  Condition
C1 follows from property~(1).  Conditions C3, C4 and C5 follow from property
(2) and the assumption that $F$ is bipartite. Condition C2 follows from
property (3).  Hence $C$ is a certifying path for $x_1$ in~$F$.
\end{proof}

Indeed, consider the certifying path $C$ of Example~\ref{exa:aux} which
gives rise to the sequence of values $v_5,v_4,v_2,v_1$. This sequence is
a directed path in $H_F^{-v_5}$, however $v_4,v_2,v_1$ is not a directed
path in $H_F^{-v_4}$, $v_2,v_1$ is not a directed path in $H_F^{-v_2}$,
and $v_1$ is not a directed path in $H_F^{-v_1}$.

Lemma \ref{path2} suggests a simple strategy for finding a certifying
path for $x_1$ in $F$, if one exists.  For each value $v$ we search for
a directed path $v_k,\dots,v_1$ from $v=v_k$ to $v_1=\eta(x_1)$ in
$H_F^{-v}$.  If we find such a path $v_k,\dots,v_1$, we check for each
subsequence $v_i,\dots,v_1$, $1\leq i < k$, whether it is a directed
path in $H_F^{-v_i}$. If the answer is \textsc{no} for all $i$, then
$v_k,\dots,v_1$ satisfies the conditions of Lemma~\ref{path2}.  Hence
the sequence of arguments in $X$ whose values form $P$ is a certifying
path for $x_1$ in $F$.  If, however, the answer is \textsc{yes} for some
$i<k$, we take the smallest $i$ for which the answer is \textsc{yes}.  Now
the sequence $v_i,\dots,v_1$ satisfies the conditions of
Lemma~\ref{path2} and so gives rise to a certifying path for $x_1$ in
$F$.  On the other hand, if there is no value $v$ such that $H_F^{-v}$
contains a directed path from $v$ to $v_1$, then there is no certifying
path for $x_1$ in $F$.
The pseudo code for this algorithm is given in Figure~\ref{fig:algo}.

\begin{figure}[tbh]
\centering
\begin{minipage}{0.8\linewidth}
    \hrule height.35mm 

\medskip
\sffamily
\begin{tabbing}
  \quad \=\quad \=\quad\=\quad \=\quad \=\quad \kill
  Algorithm {\sc Detect Certifying Path}\\[2pt]
  Input: {\normalfont value-based system $F=(X,A,V,\eta)$, query argument $x_1\in X$}\\[2pt]
  Output: {\normalfont a directed path in $H_F$ that corresponds to a  certifying path
    for $x_1$ in $F$,}\\
  {\normalfont or \textsc{no} if there is no certifying path for $x_1$ in $F$}\\[6pt]
    for all $v\in V$ do\\[2pt]
    \> \>   check if $H_F^{-v}$ contains a directed path from $v$ to $v_1$\\[2pt]
    \> \>   if yes do\\[2pt]
    \> \> \> \> find such a path,  $v_k,\dots,v_1$,  $v=v_k$\\[2pt]
    \> \> \> \> for $i=1,\dots,k$ do\\[2pt]
    \> \>  \> \> \> check if $v_i,\dots,v_1$ is a directed path in $H_F^{-v_i}$\\[2pt]
    \>  \>  \> \> \> if yes, output $v_i,\dots,v_1$ and terminate\\[2pt]
    return \textsc{no} and terminate
\end{tabbing}
    \hrule height.35mm 
\end{minipage}
\caption{Polynomial-time algorithm for the detection of a certifying
  path in a bipartite value-based system of value-width 2.}  \label{fig:algo}
\vspace{-10pt}
\end{figure}

\begin{PRO}\label{treealgo}
  The algorithm {\sc Detect Certifying Path} correctly returns a
  certifying path for $x_1$ in $F=(X,A,V,\eta)$ if one exists and
  returns NO otherwise in time $O(|V|^2\cdot (|X|+|A|+|V|))$.
\end{PRO}
\begin{proof}
  The correctness of {\sc Detect Certifying Path} follows from Lemma
  \ref{path2}. For $v\in V$, building $H_F^{-v}$ and finding a shortest
  directed path from $v$ to $v_1=\eta(x_1)$, if one exists, takes linear
  time in the input size of $F$ (which we estimate by the term
  $O(|X|+|A|+|V|)$).  As we iterate over all vertices of $V$, and we
  check for at most $\Card{V}$ subsequences $v_i,\dots,v_1$ whether it
  is a directed path in $H_F^{-v_i}$, the claimed running time follows.
\end{proof}

We are now ready to combine the above results to a proof of
Theorem~\ref{bipeasy}.  Statement (A) of the theorem follows from
Lemma~\ref{certpathsa} and Proposition~\ref{treealgo}.  Statement (B)
follows from Statement (A) and Lemma~\ref{certpathoa}.

\section{Linear-time algorithm for value-based systems of bounded treewidth}\label{sec:tw}

As mentioned above, it is known that both acceptance problems remain
intractable for value-based systems whose graph structure is a tree.
This is perhaps not surprising since two arguments can be considered as
linked to each other if they share the same value. In fact, such links
may form cycles in an otherwise tree-shaped value-based system.
Therefore we propose to consider the \emph{extended graph structure} of
the value-based system (recall Definition~\ref{def:graphs} in
Section~\ref{subsection:graphs}) that takes such links into account. We
show that the problems \textsc{Subjective} and \textsc{Objective
  Acceptance} are easy for value-based systems whose extended graph
structure is a tree, and more generally, the problems can be solved in
\emph{linear-time} for value-based systems with an extended graph
structure of \emph{bounded treewidth}.

Treewidth is a popular graph parameter that indicates in a certain sense
how similar a graph is to a tree.  Many otherwise intractable graph
problems (such as \textsc{3-Colorability}) become tractable for graphs
of bounded treewidth. Bounded treewidth (and related concepts like
\emph{induced width} and \emph{d-tree width}) have been successfully
applied in many areas of AI, see, e.g.,
\cite{Freuder85,Dechter99,Darwiche01,GottlobPichlerWei06}.  Deciding
acceptance for argumentation frameworks of bounded treewidth has been
investigated by Dunne~\cite{Dunne07} and by Dvor{\'a}k, Pichler, and
Woltran~\cite{DvorakPichlerWoltran10}.  However, for value-based
argumentation, the concept of bounded tree-width has not been applied
successfully: the basic decision problems for value-based systems remain
intractable for value-based systems of value width~3 whose graph
structure has treewidth~1~\cite{Dunne07}. Hardness even prevails for value-based systems
whose value graph has pathwidth~2~\cite{Dunne10}.  These negative results are
contrasted by our Theorem~\ref{mso}, which indicates that the extended
graph structure seems to be a suitable and adequate graphical model for
value-based systems.

\medskip\noindent The treewidth of a graph is defined using the
following notion of a tree decomposition (see,
e.g.,~\cite{Bodlaender93b}).
\begin{DEF}
 A \emph{tree decomposition} of an (undirected) graph 
$G=(V,E)$ is a pair $(T,\chi)$ where $T$ is a tree and $\chi$ is a labeling
function that assigns each tree node $t$ a set $\chi(t)$ of vertices of
the graph $G$  such that the
following conditions hold: 
\begin{enumerate}
\item~Every vertex of $G$ occurs in $\chi(t)$ for
some tree node~$t$. 
\item~For every edge $\{u,v\}$ of $G$ there is a tree node
$t$ such that $u,v\in \chi(t)$.
\item~For every vertex $v$ of $G$, the tree nodes $t$ with $v\in
  \chi(t)$ form a connected subtree of~$T$.
\end{enumerate}
 The
\emph{width} of a tree decomposition $(T,\chi)$ is the size of a largest bag
$\chi(t)$ minus~$1$ among all nodes~$t$ of~$T$.  A tree decomposition of
smallest width is \emph{optimal}.  The \emph{treewidth} of a graph $G$ is the
width of an optimal tree decomposition of~$G$. 
\end{DEF}

\begin{figure}[tbh]
\centering
  \begin{tikzpicture}[scale=0.8]
    \tikzstyle{every node}=[circle,minimum size=6mm,inner sep=0pt]
    \draw
    (.75,-6) node () {$G_F^\ext$} 
    (0,0)   node[draw] (A) {$a$}  
    (1.5,0) node[draw] (B) {$b$} 
    (0,-2)   node[draw] (C) {$c$} 
    (1.5,-2) node[draw] (D) {$d$} 
    (1.5,-4) node[draw] (F) {$f$} 
    (0,-4)   node[draw] (E) {$e$}

    ;
    \draw[thick]

    (B) edge (A)
    (C) edge (D)
    (D) edge (B)
    (A) edge (D)
    (F) edge (C)
    (A) edge[bend right] (E)
    (E) edge (F)
    ;
    \tikzstyle{every node}=[ellipse,minimum width=14mm]
    \draw
    (8+.75,-6) node () {$(T,\chi)$} 
    (8+0,0) node[draw] (1) {$a,b,d$}
    (8+.5,-1.5) node[draw] (2) {$a,d,c$}
    (8-1,-3) node[draw] (3) {$d,c$}
    (8+1.5,-3) node[draw] (4) {$a,c,f$}
    (8+2,-4.5) node[draw] (5) {$a,f,e$}
;
\draw[thick]
    (1)--(2)--(4)--(5) (2)--(3)
;

\end{tikzpicture} 
\caption{A graph and its tree
  decomposition.}\label{fig:treedecomp}
\end{figure}
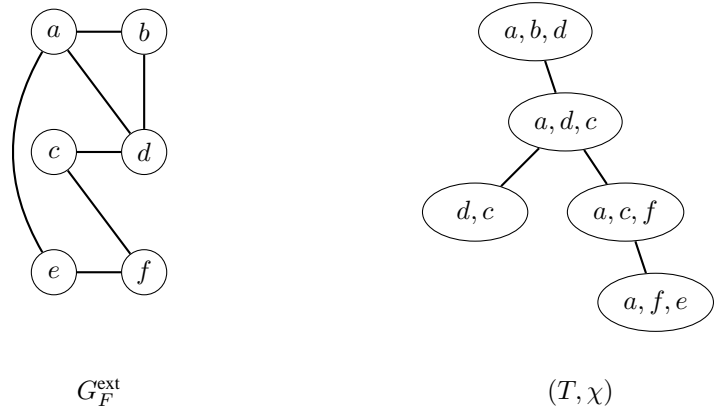

\begin{EXA}
  Figure~\ref{fig:treedecomp} exhibits a graph (the extended graph
  structure of the value-based system of Example~\ref{exa:vaf}) and a
  tree decomposition of it. The width of the tree decomposition is $2$,
  and it is not difficult to see that this is optimal. Hence the
  treewidth of the graph in the figure is~2.
\end{EXA}

\subsection{Linear-time tractability for value-based systems with
  extended graph structures of bounded treewidth}

We are going to establish the following result.

\begin{THE}\label{mso}
  The problems \textsc{Subjective} and \textsc{Objective Acceptance} can
  be decided in linear time for value-based systems whose extended graph
  structure has bounded treewidth.
\end{THE}

To achieve tractability we have to pay a price in generality: The
mentioned hardness results of \cite{Dunne10,Dunne07} imply that if
\textsc{Subjective Acceptance} is fixed-parameter tractable for any
parameter~$p$, then, unless $\P=\NP$, parameter $p$ cannot be bounded by
a function of any of the following three parameters: the treewidth of the graph
structure, the treewidth of the value graph, and the value-width. This
even holds if the bounding function is exponential.  Indeed, the
treewidth of the extended graph structure can be arbitrarily large for
value-based systems where one of these three parameters is bounded by
a constant.

The reminder of this section is devoted to a proof of Theorem~\ref{mso}. We
shall take a logic approach and use the celebrated result of
Courcelle~\cite{Courcelle87}, which states that all properties that can
be expressed in a certain formalism (Monadic Second-Order logic, MSO)
can be checked in linear time for graphs (or more generally, for finite
structures) of bounded treewidth. Courcelle's Theorem is constructive in
the sense that it not only promises the existence of an algorithm for
the particular problem under consideration, but it provides the means
for actually producing such an algorithm.  The algorithm produced in
this general and generic way leaves much room for improvement and
provides the basis for the development of problem-specific and more
practical algorithms.

In the following we use Courcelle's result as laid out by Flume and
Grohe~\cite{FlumGrohe06}.  Let $S$ denote a finite relational structure
and $\phi$ a sentence in monadic second-order logic (MSO logic)
on~$S$. That is, $\phi$ may contain quantification over atoms (elements
of the universe) and over sets of atoms.  Furthermore, we associate with
the structure $S$ its~\emph{Gaifman graph} $G(S)$, whose vertices are
the atoms of $S$, and where two distinct vertices are joined by an edge
if and only if they occur together in some tuple of a relation
of~$S$. We define the \emph{treewidth of structure $S$} as the treewidth
of its Gaifman graph $G(S)$. Now Courcelle's theorem states that for a
fixed MSO sentence $\phi$ and a fixed integer~$k$, one can check in
linear time whether $\phi$ holds for a given relational structure of treewidth
at most~$k$. 
The proof of Theorem~\ref{mso} boils down to the following
two tasks:

\begin{description}
\item{\emph{Task A.}} To represent a value-based system $F$ and a query
  argument $x_1$ by a relational structure $S[F,x_1]$ such that bounded
  treewidth of the extended graph structure of $F$ implies bounded
  treewidth of $S[F,x_1]$.
\item{\emph{Task B.}} To construct formulas $\phi_s$ and $\phi_o$ in MSO
  logic such that for every value-based system $F$ and every argument
  $x_1$ of $F$ it holds that $\phi_s$ is true for $S[F,x_1]$ if and only
  if $x_1$ is subjectively accepted in~$F$, and $\phi_o$ is true for
  $S[F,x_1]$ if and only if $x_1$ is objectively accepted in $F$.
\end{description}

\subsection{Reference graphs}
For many problems it is rather straight-forward to find an MSO
formulation so that Courcelle's Theorem can be applied. In our case,
however, we have to face the difficulty that we have to express that ``a
certain property holds for some total ordering'' (subjective acceptance)
and ``a certain property holds for all total orderings'' (objective
acceptance), which cannot be directly expressed in MSO. Our solution to
this problem lies in the introduction of an auxiliary directed graph
$R$, the \emph{reference graph}, which will allow us to quantify over
total orderings of $V$.  The relational structure $S[F,x_1]$ will then
be defined to represent $F$ together with $R$.

\begin{DEF}
  Let $F=(X,A,V,\eta)$ be a value-based system and let $\prec$ be an
  arbitrary but fixed total ordering of $V$.  The \emph{reference graph}
  $R=(V,E_R)$ is the directed graph where $V$ is the set of values of
  $F$ and $E_R$ consists of all directed edges $(u,v)$ for which
  \begin{enumerate}
  \item $u \prec v$ in the fixed ordering, and
  \item $A$ contains an attack $(x,x')$ with $\eta(x)=u$ and
    $\eta(x')=v$ or $\eta(x)=v$ and $\eta(x')=u$.
  \end{enumerate}

  For a subset $Q \subseteq E_R$ let $R[Q]=(V,E_R[Q])$ be the directed
  graph obtained from the reference graph $R$ by reversing all edges in
  $Q$, i.e., $E_R[Q]:=\SB (u,v) \SM (u,v) \in E_R \setminus Q) \SE \cup
  \SB (v,u) \SM (u,v) \in E_R \cap Q \SE$.

  We also define the abstract argumentation system $F[Q]:=(X,A[Q])$ as
  the system obtained from $F$ with $A[Q]:=\SB (u,v) \in A \SM
  (\eta(u),\eta(v)) \notin E_R[Q]) \SE$.
\end{DEF}

Note that the reference graph $R$ is by definition acyclic (in contrast
to the value graph $G_F^\val$ whose definition is similar but distinct).

Every specific audience $\leq$ of $F$ can now be represented by some
subset $Q \subseteq E_R$ for which the directed graph $R[Q]$ is acyclic,
and conversely, every set $Q \subseteq E_R$ such that $R[Q]$ is acyclic
represents a specific audience $\leq$. These observations are made
precise in the following lemma whose easy proof is omitted.

\begin{LEM}\label{lem:Q}
  An argument $x_1$ is subjectively accepted in $F$ if and only if there
  exists a set $Q \subseteq E_R$ such that $R[Q]$ is acyclic and $x_1$ is in
  the unique preferred extension of $F[Q]$.  An argument $x_1$ is
  objectively accepted in $F$ if and only if for every set $Q \subseteq E_R$
  such that $R[Q]$ is acyclic it holds that $x_1$ is in the unique preferred
  extension of $F[Q]$.
\end{LEM}

Since we can test for acyclicity with MSO logic (see the next
subsection), we can now express subjective and objective acceptance in
MSO logic as ``a certain property holds for some subset $Q$ of $E_R$ for
which $R[Q]$ is acyclic'' and ``a certain property holds for all subsets
$Q$ of $E_R$ for which $R[Q]$ is acyclic'', respectively.  Next we give
a more detailed description of how to accomplish the two tasks for our
proof.

\subsection{Task A: representing the value-based system}
We define a relational structure $S[F,x_1]$ that represents the
value-based system $F$ together with the reference graph~$R=(V,E_R)$.
The universe of $S[F,x_1]$ is the union of the sets $X$, $V$, and $E_R$.
$S[F,x_1]$ has one unary relation $U_a^*$ and four binary relations $H$,
$T$, $B_a$ and $B_\eta$ that are defined as follows:
\begin{enumerate}
\item $\text{U}_a^*(x)$ if and only if $x=x_1$ (used to ``mark'' the query
  argument).
\item $\text{T}(t,(u,v))$ if and only if $t=u$ (used to represent the
  ``tail relation'' of $E_R$)
\item $\text{H}(h,(u,v))$ if and only if $h=v$ (used to represent the
  ``head relation'' of $E_R$)
\item $\text{B}_a(x,y)$ if and only if $(x,y)\in A$ (used to represent the
  attack relation).
\item $\text{B}_\eta(x,v)$ if and only if $\eta(x)=v$ (used to represent the
  mapping $\eta$).
\end{enumerate}

\begin{figure}[tbh]
\centering
  \begin{tikzpicture}[xscale=0.6, yscale=0.8]
    \small
    \tikzstyle{every node}=[circle,minimum size=6mm,inner sep=0pt]
    \draw  (0.75,-6.25) coordinate () {};
    \draw
    (0,0)   node[draw] (A) {$a$}  
    (1.5,0) node[draw] (B) {$b$} 
    (0,-2)   node[draw] (C) {$c$} 
    (1.5,-2) node[draw] (D) {$d$} 
    (1.5,-4) node[draw] (F) {$f$} 
    (0,-4)   node[draw] (E) {$e$}
    (.75,-5.5) node () {$F$} 
    ;
    \draw[thick,->,shorten >=1pt,>=stealth]

    (B) edge (A)
    (C) edge (D)
    (D) edge (B)
    (A) edge (D)
    (F) edge (C)
    (A) edge[bend right] (E)
    ;
    \draw  (0.75,0) ellipse (1.95 and 0.75);
    \draw  (0.75,-2) ellipse (1.95 and 0.75);
    \draw  (0.75,-4) ellipse (1.95 and 0.75);
    \draw  (-1.7,0) node () {$S$:} ;
    \draw  (-1.7,-2) node () {$E$:} ;
    \draw  (-1.7,-4) node () {$T$:} ;


    \draw
    (8+0,0)   node[draw] (A) {$a$} 
    (8+1.5,0) node[draw] (B) {$b$} 
    (8+0,-2)   node[draw] (C) {$c$} 
    (8+1.5,-2) node[draw] (D) {$d$} 
    (8+1.5,-4) node[draw] (F) {$f$} 
    (8+0,-4)   node[draw] (E) {$e$}
    (8+3.5,0) node[rectangle,draw] (s) {$S$} 
    (8+3.5,-2) node[rectangle,draw] (e) {$E$} 
    (8+3.5,-4) node[rectangle,draw] (t) {$T$} 

    (8+4,-1) node[rectangle,draw,inner sep=2pt] (se) {$(S,E)$} 
    (8+4,-3) node[rectangle,draw,inner sep=2pt] (et) {$(E,T)$} 
    (8+6,-2) node[rectangle,draw,inner sep=2pt] (st) {$(S,T)$} 
    (8+2,-5.5) node () {$G(S[F,x_1])$} 
    ;
    \draw[thick]
    (B) edge (A)
    (C) edge (D)
    (D) edge (B)
    (A) edge (D)
    (F) edge (C)
    (A) edge[bend right] (E)
    
    (A) edge[bend left] (s)
    (B) edge (s)
    (C) edge[bend left] (e)
    (D) edge (e)
    (E) edge[bend left] (t)
    (F) edge (t)

    (s) edge (se)
    (se) edge (e)
    (e) edge (et)
    (et) edge (t)
    (s) edge[bend left] (st)
    (st) edge[bend left] (t)

    ;
  \end{tikzpicture}%
\caption{Value-based system $F$ of Example~\ref{exa:vaf} and the
  corresponding Gaifmann graph $G(S[F,x_1])$.} \label{fig:geifmann}
\end{figure}
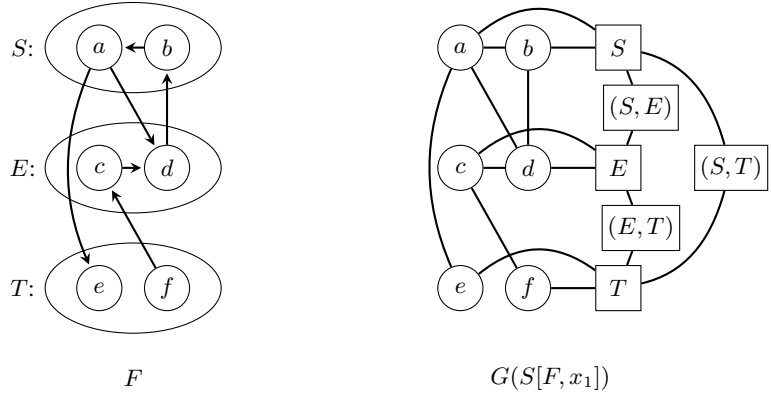

Consequently, the Gaifman graph of $S[F,x_1]$ is the graph
$G(S[F,x_1])=(V_{S[F,x_1]},E_{S[F,x_1]})$ with $V_{S_F}=X \cup V \cup
E_R$ and $E_{S_F} = \SB \{u,v\} \SM (u,v) \in \text{T} \cup \text{H}
\cup \text{B}_a \cup \text{B}_\eta \SE$, see Figure~\ref{fig:geifmann}
for an illustration.

\begin{LEM}\label{lem:tw}
  The treewidth of $S[F,x_1]$ is at most twice the treewidth of the extended
  graphs structure of $F$ plus~$1$.
\end{LEM}
The easy proof is given in the appendix.


\subsection{Task B: expressing acceptance in MSO}

In order to define $\phi_s$ and $\phi_o$ we introduce the following
auxiliary formulas:

A formula $\text{TH}(t,h,a)$ that holds if and only if $t$ is the tail
and $h$ is the head of $a\in E_R$:
\[
\text{TH}(t,h,a) := \text{T}(t,a) \land \text{H}(h,a)
\]
A formula $\text{E}(t,h,Q)$ that holds if and only if the directed edge
$(t,h)$ is contained in $R[Q]$:
\[
\text{E}(t,h,Q) := \exists a\; [(\lnot Qa \land
  \text{TH}(t,h,a)) \lor (Qa \land \text{TH}(h,t,a))]
\]
A formula $\text{ACYC}(Q)$ that checks whether $R[Q]$ is acyclic. We use
the well-known fact that a directed graph contains a directed cycle if
and only if there is a nonempty set $C$ of vertices each having an
out-neighbor in~$C$ (see, e.g., \cite{BangjensenGutin00}).
\[
\text{ACYC}(Q) := \lnot \exists C\; ( 
  \exists x Cx \land
  \forall t \exists h [Ct
  \rightarrow (Ch \land \text{E}(t,h,Q))])
\]
A formula $\text{B}_a'(t,h,Q)$ that holds if and only if $t$ attacks $h$
in $F[Q]$:
\[ 
\text{B}_a'(t,h,Q) := \text{B}_a(t,h) \land \exists v_h \exists v_t \; [
\text{B}_\eta(t,v_t) \land \text{B}_\eta(h,v_h) \land \lnot
\text{E}(v_h,v_t,Q) ]
\] 
A formula $\text{ADM}(S,Q)$ that checks whether a set $S \subseteq X$ is
admissible in $F[Q]$:
\[
\text{ADM}(S,Q) := \forall x \forall y \; [(\text{B}_a'(x,y,Q) \land Sy
) \rightarrow (\lnot Sx \land \exists z ( Sz \land
\text{B}_a'(z,x,Q)))]
\] 
Now the formula $\phi_s$ can be defined as follows:
\[
\phi_s := \exists Q \;[ \text{ACYC}(Q) \land (\exists S (\forall x
(\text{U}_a^*(x) \rightarrow Sx) \land \text{ADM}(S,Q)))]\]

It follows from Lemma~\ref{lem:Q} that $\phi_s$ is
true for $S[F,x_1]$ if and only if $x_1$ is subjectively accepted in $F$.  A
trivial modification of $\phi_s$ gives us the desired sentence $\phi_o$:
\[
\phi_o := \forall Q \;[ 
\text{ACYC}(Q) \rightarrow (\exists S (\forall x (\text{U}_a^*(x)
\rightarrow Sx) \land \text{ADM}(S,Q)))]
\]
It follows from Lemma~\ref{lem:Q} that $\phi_o$ is true for $S[F,x_1]$ if and
only if $x_1$ is objectively accepted in $F$.

We summarize the above construction in the next lemma.
\begin{LEM}\label{lem:formula}
  There exists an MSO sentence $\phi_s$ such that $\phi_s$ is true for
  $S[F,x_1]$ if and only if $x_1$ is subjectively accepted in
  $F$. Similarly, there exists an MSO sentence $\phi_o$ such that
  $\phi_o$ is true for $S[F,x_1]$ if and only if $x_1$ is objectively
  accepted in $F$.
\end{LEM}

In view of Lemmas~\ref{lem:tw} and \ref{lem:formula},
Theorem~\ref{mso} now follows by Courcelle's Theorem.

\bigskip\noindent
If both the treewidth of the value graph and the value-width of an value-based system
are bounded, then also the extended graph structure has bounded
treewidth, hence we have the following corollary.

\begin{COR}
  The problems \textsc{Subjective} and \textsc{Objective Acceptance} can
  be decided in linear time for value-based systems for which both
  value-width and the treewidth of their value graphs are bounded.
\end{COR}
\begin{proof}
  Let $k$ and $k'$ be constants.  Let $(T,\chi)$ be a tree decomposition
  of the value graph of a value-based system $F$ of width $k$ and assume
  the value-width of $F$ is $k'$.  Then $(T,\chi')$ with
  $\chi'(t)=\bigcup_{v \in \chi(t)} \eta^{-1}(v)$ is a tree
  decomposition of the extended graph structure of $F$.  Since
  $\Card{\chi'(t)}\leq \Card{\chi(t)} \cdot k' \leq (k+1)k'$ holds for
  all nodes $t$ of $T$, it follows that the width of $(T,\chi')$ is
  bounded by the constant $k''=(k+1)k'-1$.  We conclude, in view of
  Theorem~\ref{mso}, that we can decide both acceptance problems for $F$
  in linear time.
\end{proof}

\section{Conclusion}\label{sec:concl}

We have studied the computational complexity of persuasive argumentation
for value-based argumentation frameworks under structural restrictions.
We have established the intractability of deciding subjective or
objective acceptance for value-based systems with value-width~2 and
attack-width~1, disproving conjectures stated by Dunne. It might be
interesting to note that our reductions show that intractability even
holds if the attack relation of the value-based system under
consideration forms a directed acyclic graph. On the positive side we
have shown that value-based systems with value-width~2 whose graph
structure is bipartite are solvable in polynomial time. These results
establish a sharp boundary between tractability and intractability of
persuasive argumentation for value-based systems with value-width~2.
Furthermore we have introduced the notion of the \emph{extended graph
  structure} of a value-based system and have shown that subjective and
objective acceptance can be decided in linear-time if the treewidth of
the extended graph structure is bounded (that is, the problems are
\emph{fixed-parameter tractable} when parameterized by the treewidth of
the extended graph structure). This is in strong contrast to the
intractability of the problems for value-based systems where the
treewidth of the graph structure or the treewidth of their value graph
is bounded.  Therefore we conclude that the extended graph structure
seems to be an appropriate graphical model for studying the
computational complexity of persuasive argumentation. It might be
interesting for future work to extend this study to other
graph-theoretic properties or parameters of the extended graph
structure.





\section*{Appendix: Technical proofs}\label{app}
\begin{proof}[Proof of Lemma~\ref{certpathsa}]
  Let $C=(x_1,z_1,\dots,x_k,z_k,t)$ be a certifying path for $x_1$ in
  $F$. Take a specific audience $\leq$ such that $\eta(x_1) < \dots <
  \eta(x_k) < \eta(t)$ and all other values in $V$ are smaller than
  $\eta(x_1)$.  We claim that the unique preferred extension
  $P=\GE(F_\leq)$ of $F_{\leq}$ includes $\{x_1,\dots , x_k,t\}$ and
  excludes $\{z_1,\dots , z_{k}\}$, which means that $x_1$ is
  subjectively accepted in $F$. It follows from C5 that $t$ is not
  attacked by any other argument in $F_\leq$ and hence $t \in P$ (see
  also Section~\ref{sec:pre} for a description of an algorithm to find
  the unique preferred extension of an acyclic abstract argumentation
  system). From C4 it follows that $z_k\notin P$. Furthermore, if there
  exists an argument $z \neq t$, $\eta(t)=\eta(z)$ then either $(t,z)
  \in A_\leq$ or $z$ does not attack an argument in
  $\{x_1,\dots,x_k,t\}$.  In the first case $z \notin P$ and does not
  influence the membership in $P$ for any other arguments in $X$. In the
  second case $z \in P$ but it does not attack any argument in
  $\{x_1,\dots,x_k,t\}$. In both cases it follows that $x_k \in
  P$. Using C3 it follows that $z_{k-1} \notin P$ and since we already
  know that $z_k \notin P$ it follows that $x_{k-1} \in P$.  A repeated
  application of the above arguments establishes the claim, and hence
  $x_1\in P$ follows.

  Conversely, suppose that there exists a specific audience $\leq$ such
  that $x_1$ is contained in the unique preferred extension
  $P=\GE(F_\leq)$ of $F_{\leq}$.  We will now construct a
  certifying path $C$ for $x_1$ in $F$.  Clearly, if there is no $z_1
  \in X\setminus \{x_1\}$ with $\eta(z_1)=\eta(x_1)$ and $(z_1,x_1) \in
  A$, then $(x_1)$ is a certifying path for $x_1$ in $F$. Hence, it
  remains to consider the case where such a $z_1$ exists.  Since $x_1
  \in P$ it follows that $z_1 \notin P$. The sequence $(x_1,z_1)$
  clearly satisfies properties C1--C3.  We now show that we can always
  extend such a sequence until we have found a certifying path for $x_1$
  in $F$. Hence, let $S=(x_1,z_1,\dots,x_l,z_l)$ be such a sequence
  satisfying conditions C1--C3, and in addition assume $S$ satisfies the
  following two conditions:
\begin{enumerate}
  \item[S1] It holds that $\eta(x_1) < \dots < \eta(x_l)$.
  \item[S2] For every $1 \leq i \leq l$ we have $x_i \in P$ and $z_i \notin
    P$.
  \end{enumerate}
  Clearly, the sequence $(x_1,z_1)$ satisfies S1 and S2, hence we can
  include these conditions in our induction hypothesis. It remains to
  show how to extend $S$ to a certifying path. Let $Y:=\SB y\in P \SM
  (y,z_l) \in A \land \eta(y)>\eta(x_l)=\eta(z_l)\SE$. Then $Y \neq
  \emptyset$ because $z_l\notin P$ by condition S2 and the assumption
  that $P$ is a preferred extension.  
  
  For each $y\in Y$ let $C_y=(x_1,z_1,\dots,x_l,z_l,y)$.  If there is an
  argument $y\in Y$ such that $C_y$ is a certifying path for $x_1$ in
  $F$ we are done.  Hence assume there is no such $y\in Y$.

  We choose $x_{l+1}\in Y$ arbitrarily. Note that $C_{x_{l+1}}$
  satisfies the condition~C4; $(x_{l+1},z_l)\in A$ (as $x_{l+1}\in Y$)
  and $(x_{l+1},x_i)\notin A$ for $1\leq i \leq l$ (as $x_{l+1},x_i\in
  P$ and $P$ is conflict-free). Since we assume that $C_{x_{l+1}}$ is
  not a certifying path, $C_{x_{l+1}}$ must violate C5.

  It follows that there exists some argument $z_{l+1}$ with
  $\eta(z_{l+1})=\eta(x_{l+1})$ such that $(x_{l+1},z_{l+1}) \notin A$
  and $(z_{l+1},x_i) \in A$ for some $1 \leq i \leq l+1$. We conclude
  that $S'=(x_1,z_1,\dots,x_l,z_l,x_{l+1},z_{l+1})$ satisfies conditions
  C1--C3 and S1--S2. Hence, we are indeed able to extend $S$ and will
  eventually obtain a certifying path for $x_1$ in~$F$.
\end{proof}

\begin{proof}[Proof of Lemma~\ref{certpathoa}]
  Assume that $x_1$ is objectively accepted in $F$.  Suppose there is a
  $p \in X$ that attacks $x_1$ and $\eta(p)=\eta(x_1)$.  If we take a
  specific audience $\leq$ where $\eta(x_1)$ is the greatest element,
  then $x_1$ is not in the unique preferred extension of $F_\leq$, a
  contradiction to the assumption that $x_1$ is objectively
  accepted. Hence $\eta(p)\neq \eta(x_1)$ for all arguments $p\in X$
  that attack $x_1$.  Next suppose there is an argument $p\in X$ that
  attacks $x_1$ and is subjectively accepted in $F -\eta(x_1)$.  Let
  $\leq$ be a specific audience such that $p$ is in the unique preferred
  extension of $(F -\eta(x_1))_{\leq}$. We extend $\leq$ to a total
  ordering of $V$ ensuring $\eta(x_1) \leq \eta(p)$.  Clearly $x_1$ is
  not in the unique preferred extension of $F_\leq$, again a
  contradiction. Hence indeed for all $p\in X$ that attack $x_1$ we have
  $\eta(p)\neq\eta(x_1)$ and $p$ is not subjectively accepted in
  $F-\eta(x_1)$

  We establish the reverse direction by proving its counter positive.
  Assume that $x_1$ is not objectively accepted in~$F$.  We show that
  there exists some $p\in X$ that attacks $x_1$ and where either
  $\eta(p)=\eta(x_1)$ or $p$ is subjectively accepted in $F-\eta(x_1)$.
  Let $\leq$ be a specific audience of $F$ such that $x_1$ is not in the
  unique preferred extension $P=\GE(F_\leq)$ of $F_\leq$.  In
  view of the labeling procedure for finding $P$ as sketched in
  Section~\ref{sec:pre}, it follows that there exists some $p \in P$
  that attacks $x_1$ with $\eta(x_1) \leq \eta(p)$.  If $ \eta(x_1) =
  \eta(p)$ then we are done.  On the other hand, if $\eta(p) \neq
  \eta(x_1)$, then $p$ is in the unique preferred extension of
  $(F-\eta(x_1))_\leq$, and so $p$ is subjectively accepted in
  $F-\eta(x_1)$.
\end{proof}

\begin{proof}[Proof of Corollary~\ref{cor:bip}]
  We slightly modify the reduction from 3SAT as given in
  Section~\ref{subsection:hard}. Let $C_1,\dots,C_m$ be the clauses of
  the 3CNF formula $\Phi$. It is well-known that 3SAT remains $\NP$-hard
  for formulas where each clause is either positive (all three literals
  are unnegated variables) or negative (all three literals are negated
  variables), see \cite{GareyJohnson79}. Hence we may assume that for
  some $2\leq k \leq m$, $C_1,\dots,C_k$ are positive clauses and
  $C_{k+1},\dots,C_m$ are negative clauses.  Let $F$ and $F'$ be the two
  value-based systems corresponding to $\Phi$ as constructed in
  Section~\ref{subsection:hard}.  We obtain from $F$ the value-based
  system $F_B$ by adding a new pair of arguments $x_B,y_B$ with a new
  value $v_B=\eta(x_B)=\eta(y_B)$ and inserting the pair between the
  pairs $x_k^i,z_k^i$ and the pair $x_{k+1},z_{k+1}$.  That is, for
  $1\leq i \leq 3$ we replace the attacks $(x_{k+1},z_k^i)$ and
  $(z_{k+1},x_k^{i})$ with the attacks $(x_B,z_k^i)$ and
  $(z_B,x_k^{i})$, and we add the attacks $(x_{k+1},z_B)$,
  $(z_{k+1},x_B)$. By the same modification we obtain from $F'$ the
  value-based system $F'_B$.  Clearly Claims~\ref{claim:subj} and
  \ref{claim:obj} still hold for the modified value-based systems, i.e.,
  $\Phi$ is satisfiable if and only if $x_1$ is subjectively accepted in
  $F$, and $\Phi$ is satisfiable if and only if $x_1$ is not objectively
  accepted in $F'$.

  In order to establish the corollary it remains to show that the value
  graphs of $F_B$ and $F_B'$ are bipartite.

  We partition the set of arguments into two sets $X_0$ and $X_1$.
  $X_0$ contains the values $v_j$ for $j\leq k$, the value $v_B$, and
  the values $v_j^i$ for $j>k$.  $X_1$ contains the values $v_j$ for $j>
  k$, the values $v_j^i$ for $j\leq k$, and the value $v_t$. For $F_B'$, 
  $X_1$ contains also the value $v_0$. It is easy to check that there is
  no attack $(a,b)$ with $\eta(a),\eta(b)\in X_0$ or $\eta(a),\eta(b)\in
  X_1$, hence $F_B$ and $F_B'$ have bipartite value graphs.
\end{proof}

\begin{proof}[Proof of Lemma~\ref{lem:tw}]
  Let $G'$ be the graph obtained from
  $G(S[F,x_1])=(V_{S[F,x_1]},E_{S[F,x_1]})$ by replacing every path of
  the form $(t,(t,h),h)$ for $t,h\in V$ by an edge $\{t,h\}$; i.e.,
  $G'=(V',E')$ where $V'=X \cup V$ and $E'=(E_{S[F,x_1]} \cap \SB
  \{u,v\} \SM u,v \in X \cup V\SE) \cup \SB \{t,h\} \SM (t,h) \in E_R
  \SE$.  Conversely one can obtain $G(S[F,x_1])$ from $G'$ by
  subdividing all edges of the form $\{t,h\}$ for $t,h\in V$ and
  $(t,h)\in E_R$ with a vertex $(t,h)$.  However, subdividing edges does
  not change the treewidth of a graph \cite{Bodlaender93b}, hence it
  suffices to show that the treewidth of $G'$ is at most twice the
  treewidth of the extended graph structure of $F$ plus~1. Let
  $\mathcal{T}=(T,\chi)$ be a tree decomposition of the extended graph
  structure of $F$. We observe that $\mathcal{T}'=(T,\chi')$ where
  $\chi'(t)=\chi(t) \cup \SB \eta(v) \SM v \in X\cap \chi(t) \SE$ is a
  tree decomposition of $G'$ where $\Card{\chi'(t)}\leq
  2\cdot\Card{\chi(t)}$ for all nodes~$t$ of $T$; hence the width of
  $\TTT'$ is at most twice the width of $\TTT$ plus~1.
\end{proof}


\end{document}